\documentclass[runningheads]{llncs}

\usepackage[T1]{fontenc}
\usepackage{etoolbox}
\usepackage{multirow}%
\usepackage{amsmath,amssymb,amsfonts}%

\usepackage{amsthm}%
\usepackage{mathrsfs}%
\usepackage[title]{appendix}%
\usepackage{xcolor}%
\usepackage{textcomp}%
\usepackage{manyfoot}%
\usepackage{booktabs}%
\usepackage{algorithm}%
\usepackage{algorithmicx}%
\usepackage{algpseudocode}%
\usepackage{listings}%
\usepackage{multirow}
\usepackage{adjustbox}
\usepackage{caption}
\usepackage{subcaption}
\usepackage[T1]{fontenc}
\usepackage{amssymb}
\usepackage{graphicx}
\usepackage{commath}
\usepackage{color}
\usepackage{booktabs}
\usepackage{wrapfig}
\usepackage{multirow}
\usepackage{microtype}
\usepackage[dvipsnames]{xcolor}
\usepackage{hyperref}
\usepackage[capitalise]{cleveref}
\usepackage{microtype}
\usepackage{enumitem}
\setlist[itemize]{noitemsep, topsep=0pt}
\setlist[enumerate]{noitemsep, topsep=0pt}
\definecolor{darkgreen}{rgb}{0.0, 0.5, 0.0}
\newcommand{\astroicon}[1]{\adjustbox{trim=0cm 0.75mm 0cm 0cm}{\includegraphics[scale=0.02]{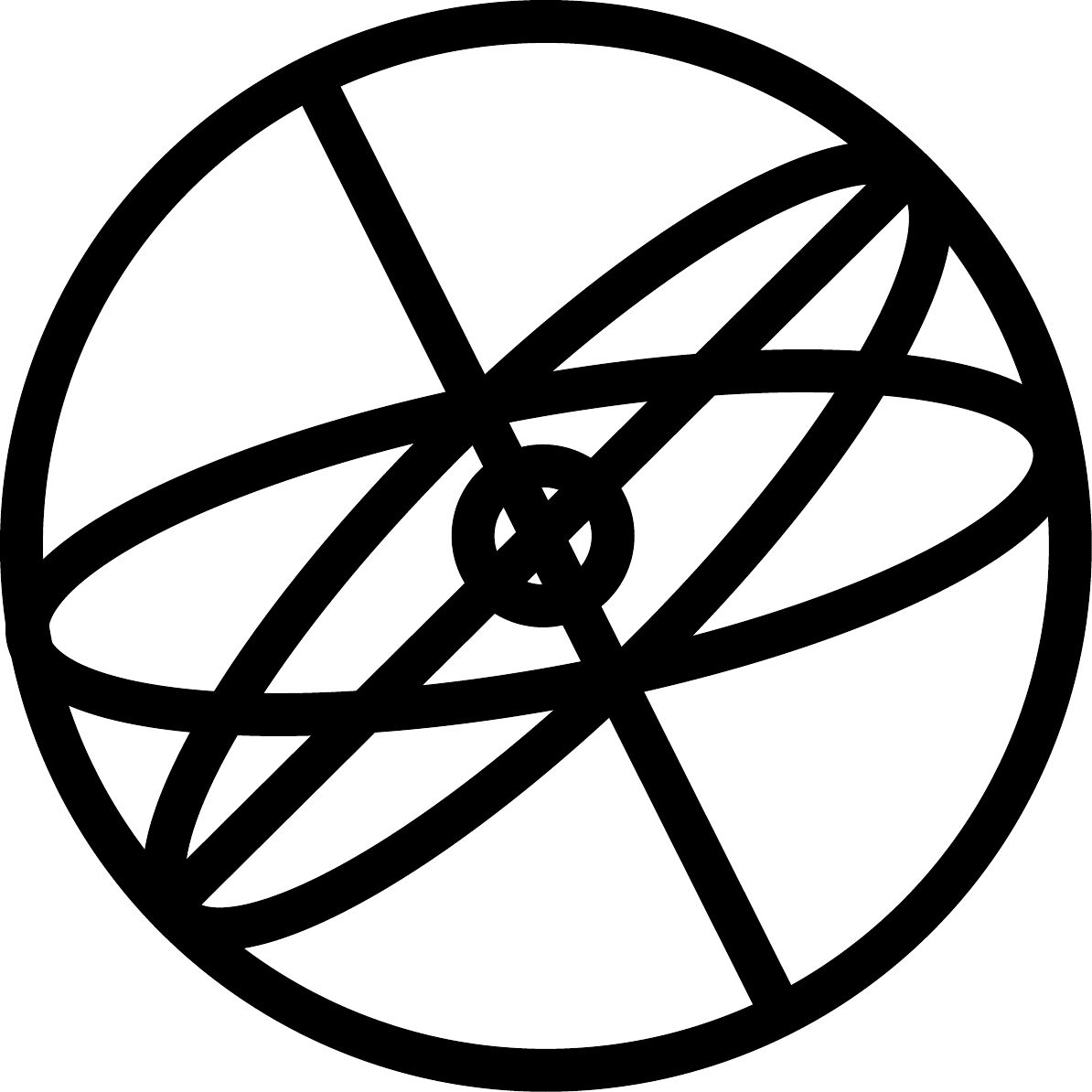}}}
\usepackage{setspace}  

\usepackage{orcidlink}
\newcommand{\blockcomment}[1]{}

\newcommand{\SB}[1]{{\textcolor{black}{#1}}}
\newcommand{\equalcontrib}{\let\thefootnote\relax\footnotetext{*equal contribution.}}
\usepackage{fontawesome5}

\begin{document}
%
\title{Text region detection in \\historical astronomical diagrams
}
\titlerunning{Text region detection in historical astronomical diagrams}

\author{Zeynep Sonat Baltac\i{}$^*$\inst{1}\orcidlink{0000−0001−6749−2407},
Rapha\"{e}l Baena$^*$\inst{1}\orcidlink{0000-0003-3214-3252},
Fei Meng\inst{1}\orcidlink{0009-0005-3840-876X},
Somkéo Norindr\inst{2},
Florence Somer\inst{2},
Matthieu Husson\inst{2},
Mathieu Aubry\inst{1}\orcidlink{0000−0002−3804−0193}}
\authorrunning{Baltac\i{} et al.}
\institute{LIGM, ENPC, IP Paris, Univ Gustave Eiffel, CNRS, Marne-la-Vallée, France \and
LTE, CNRS, PSL-Observatoire de Paris, SU, EIDA Project \\ \email{sonat.baltaci@enpc.fr}}


\maketitle              
\equalcontrib
\begin{abstract}
Text detection is a crucial task in the analysis of historical documents. While datasets and benchmarks exist for text detection in manuscripts and maps, the study of text in mathematical diagrams has received little attention. To address this, we introduce a large-scale, diverse, open-access dataset of \textbf{948 historical astronomical diagrams containing 10,940 oriented polygonal text regions}. 
Our dataset spans ten centuries (8$^{\text{th}}$ to 18$^{\text{th}}$) and seven main linguistic traditions: Arabic and Persian (115), Chinese (332), Byzantine (233), Latin (185), Hebrew (48), and Sanskrit (35). It captures a wide range of diagram styles and textual content, from symbols to multi-line paragraphs. Each text instance is annotated with ordered polygons that precisely delineate text regions and encode the reading direction. In addition, we annotated the \textbf{2,293 regions in Latin diagrams with 20 class labels}. We evaluated several strong baselines on our dataset, including TESTR,
DeepSolo++, and Poly-DETR, a simple extension of DINO-DETR that we design to predict ordered polygon vertices.
Poly-DETR achieves state-of-the-art performance on the MTHv2 and cBAD2019 benchmarks and provides a solid, simple baseline on our dataset. Code and dataset available \href{https://github.com/sonatbaltaci/textindiagrams}{online}.
\end{abstract}

\section{Introduction}
\label{sec:intro}
\begin{figure*}[ht!]
\centering
    \resizebox{0.99\linewidth}{!}{
    \begin{minipage}{\textwidth}
        \centering
    \begin{subfigure}[t]{0.45\linewidth}
    \centering
    \includegraphics[width=\linewidth]{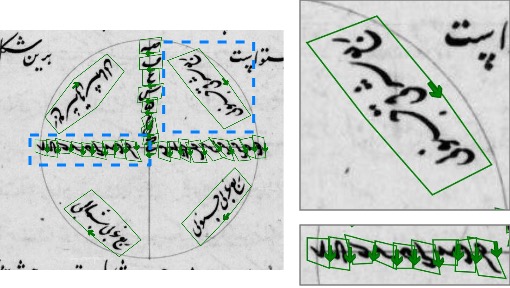}
    \caption{Annotations with reading order.\label{subfig:a}}
    \end{subfigure}
    \begin{subfigure}[t]{0.5\linewidth}
    \centering
    \includegraphics[width=0.75\linewidth]{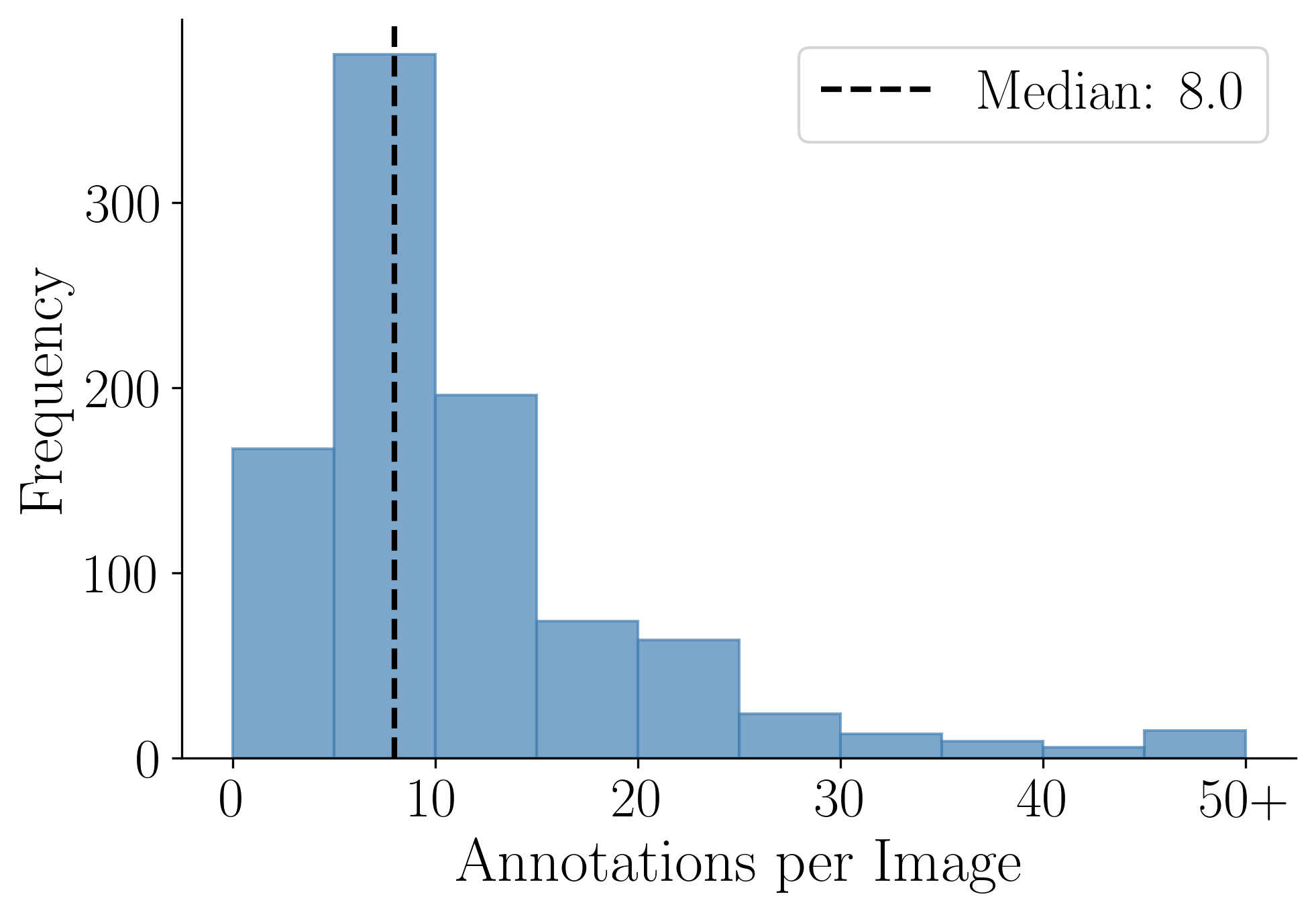}
    \caption{Text elements per diagram.\label{subfig:b}} 
    \end{subfigure}
    \begin{subfigure}[t]{0.45\linewidth}
    \centering
    \includegraphics[width=0.9\linewidth]{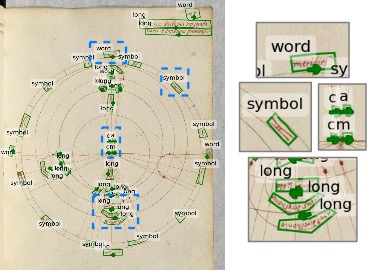}
    \caption{Text class annotations in Latin.\label{subfig:c}}
    \end{subfigure}
    \begin{subfigure}[t]{0.5\linewidth}
    \centering
    \includegraphics[width=\linewidth,trim=0 2em 0 0]{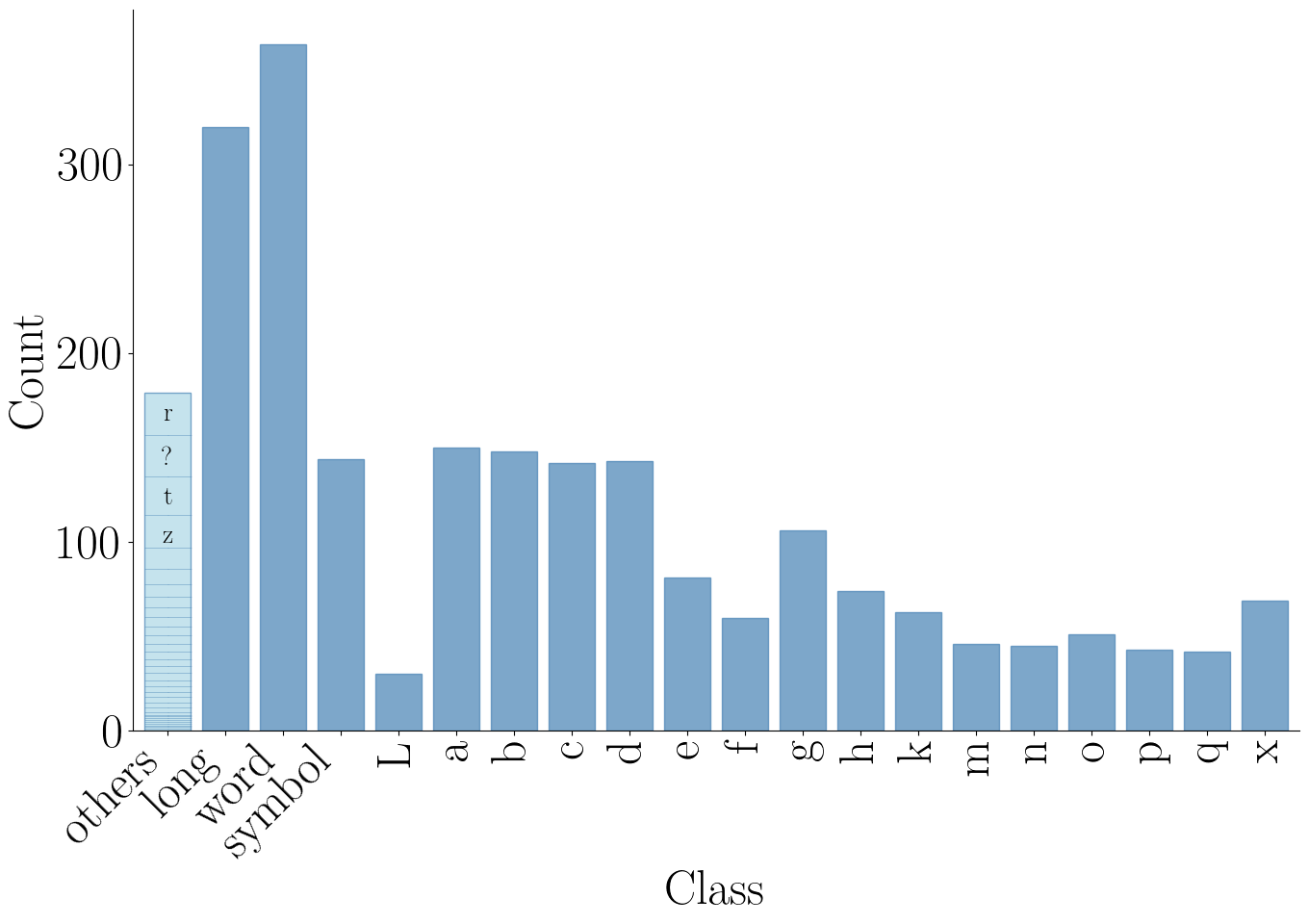}
    \caption{Class distribution of the regions in Latin.\label{subfig:d}}
    \end{subfigure}
    \\
    \subfloat[Inter-tradition diversity (Latin, Arabic, Sanskrit, Hebrew, and Byzantine).\label{subfig:e}]{
        \begin{tabular}{ccccc}
        \includegraphics[height=2.2cm]{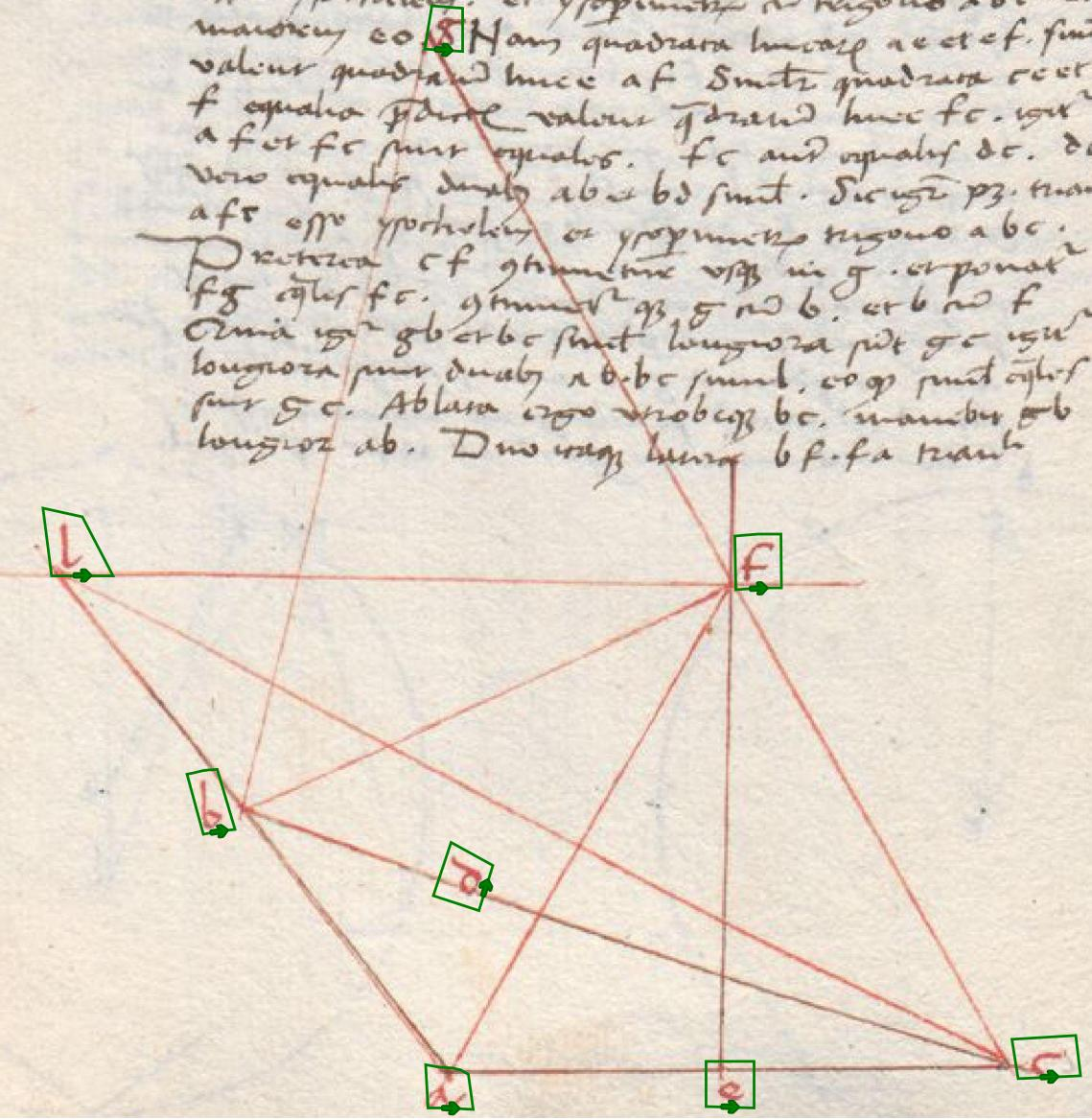} &
        \includegraphics[height=2.2cm]{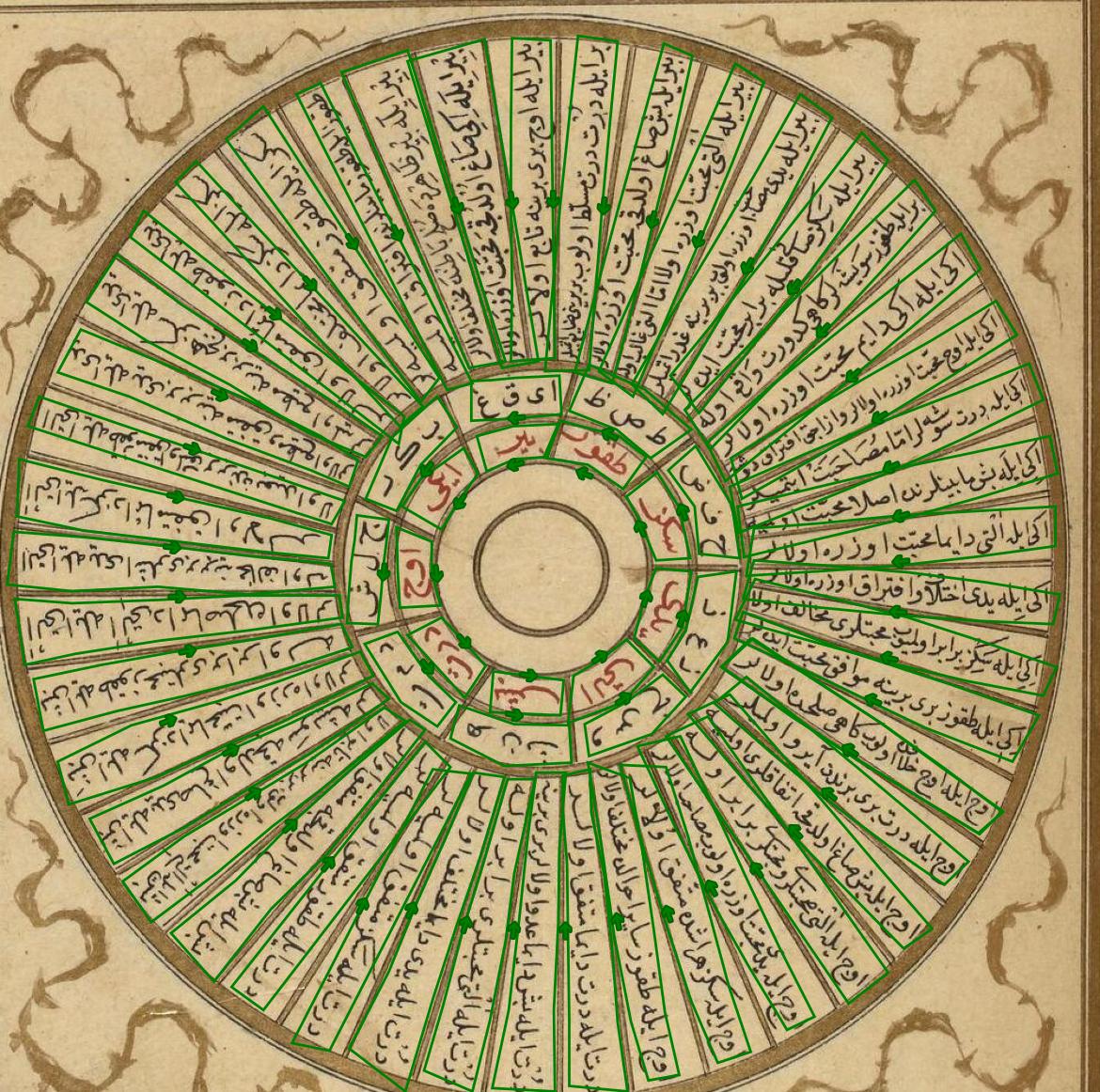} &
        \includegraphics[height=2.2cm]{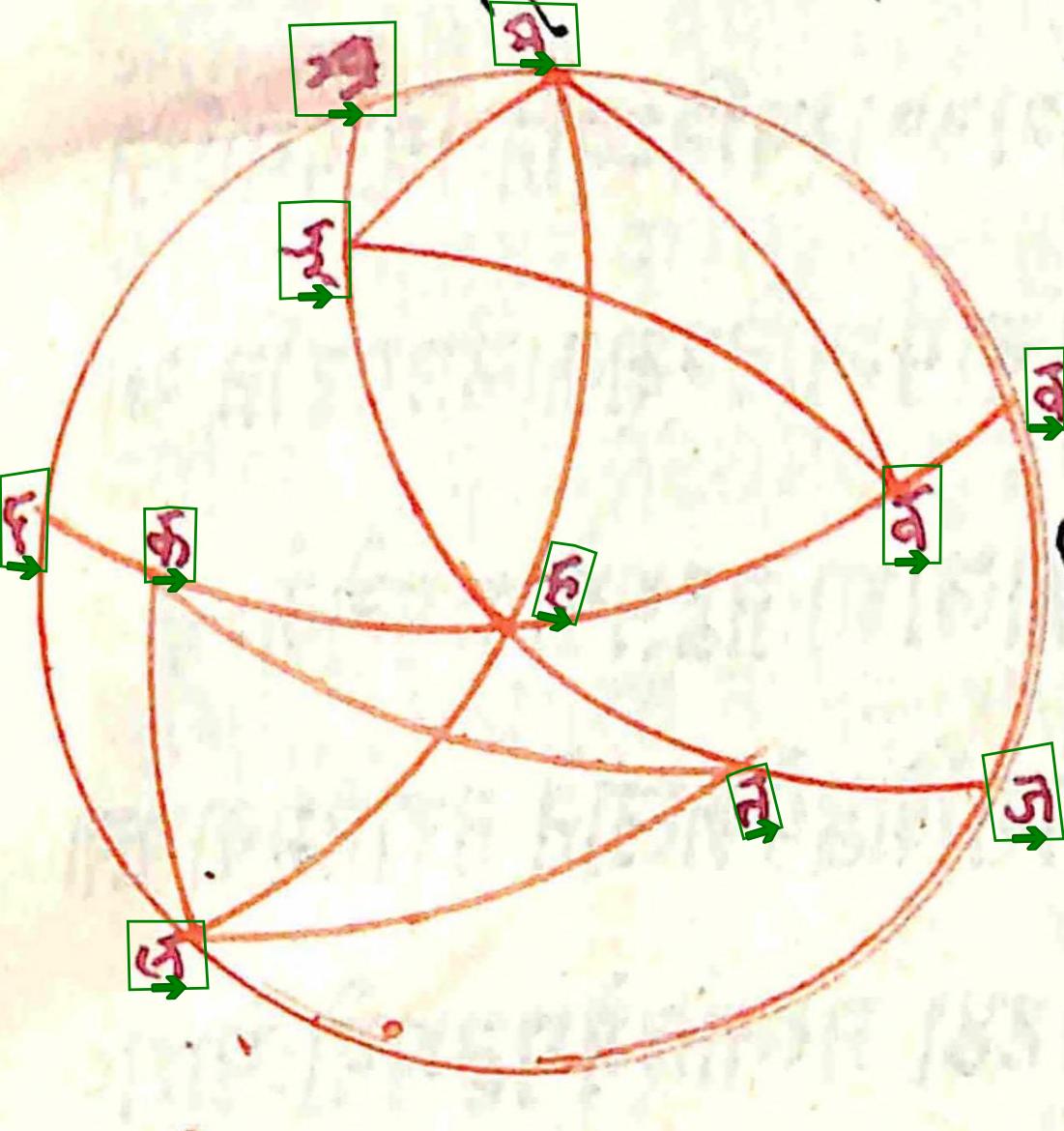} &
        \includegraphics[height=2.2cm]{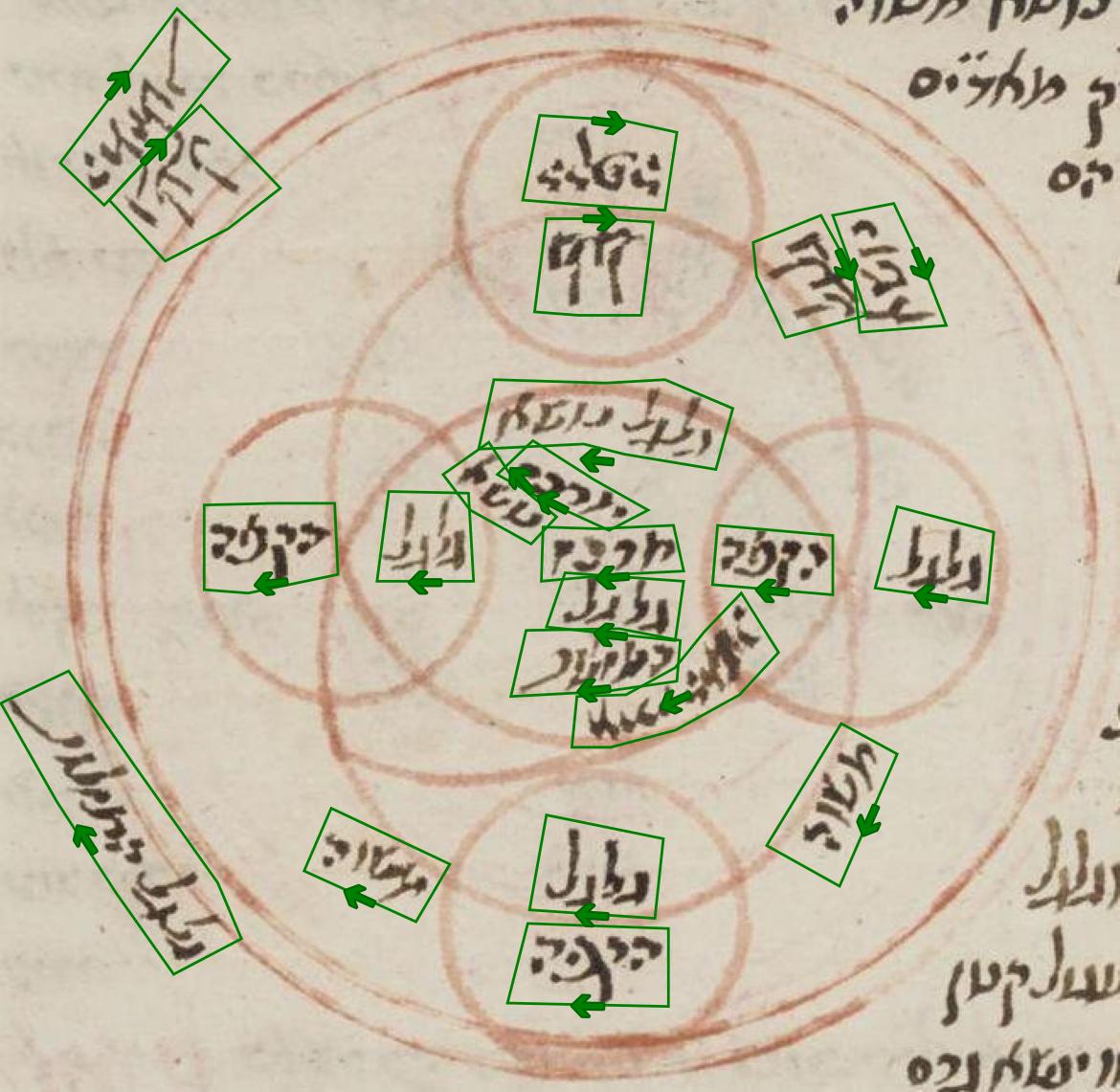} &
        \includegraphics[height=2.2cm]{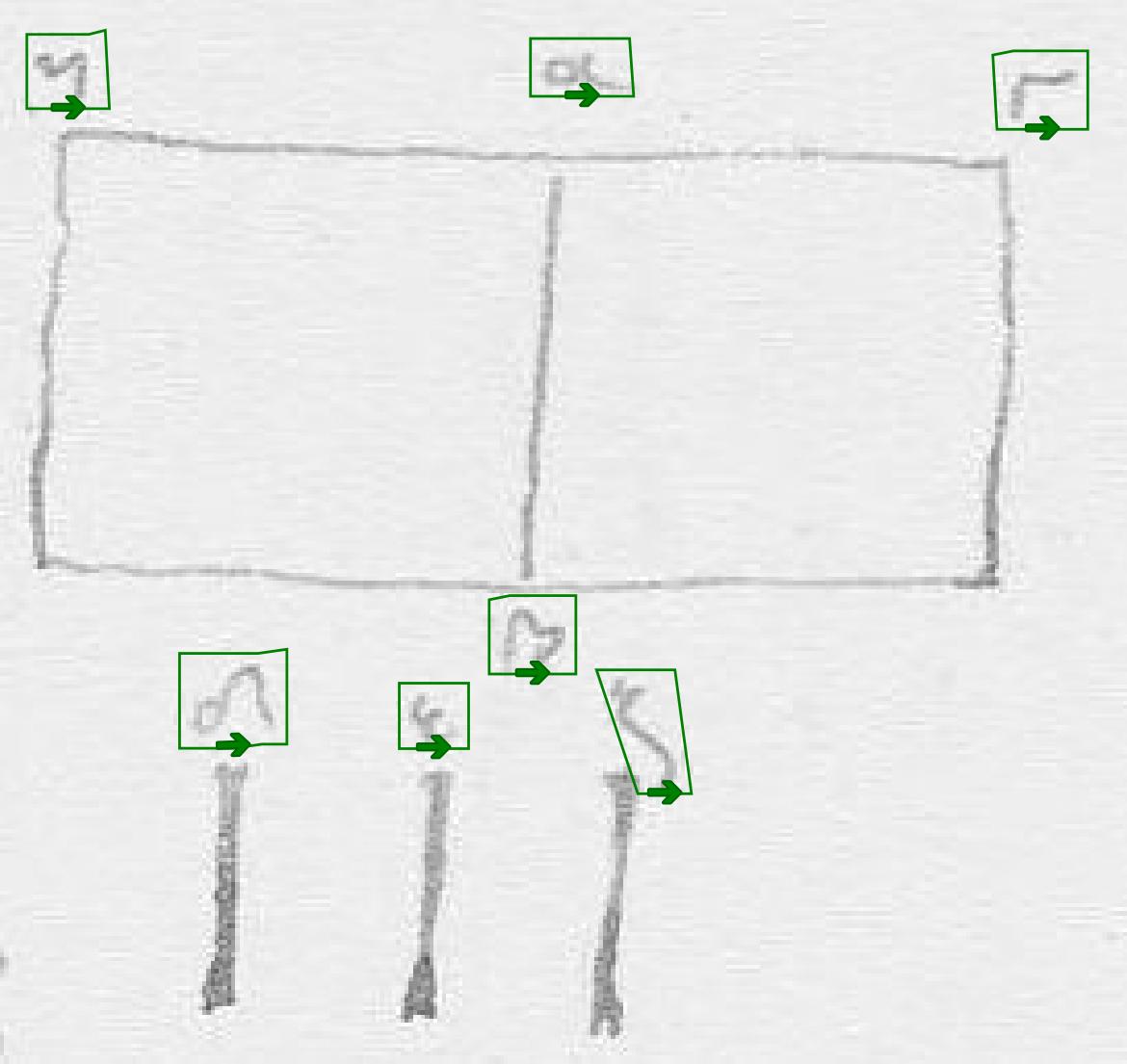} \\ 
        \end{tabular}} \\
    \subfloat[Intra-tradition diversity (Chinese).\label{subfig:f}]{
        \begin{tabular}{ccccc}
        \includegraphics[height=3cm]{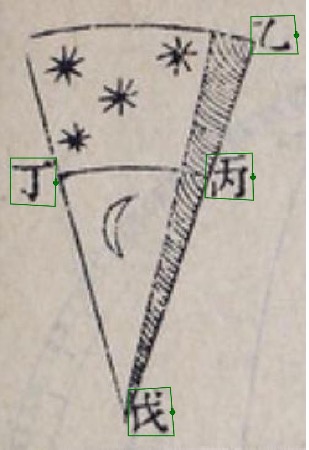} &
        \includegraphics[height=3cm]{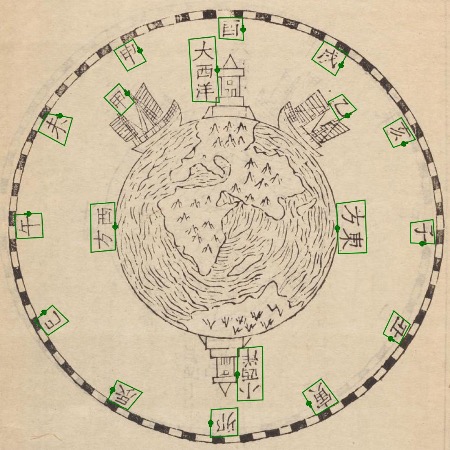} &
        \includegraphics[height=3cm]{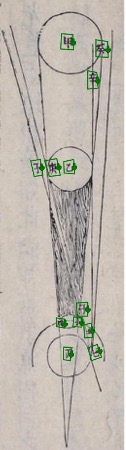} &
        \includegraphics[height=3cm]{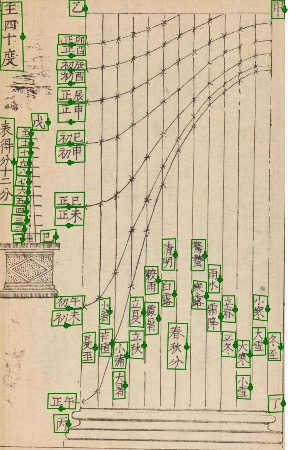} &
        \includegraphics[height=3cm]{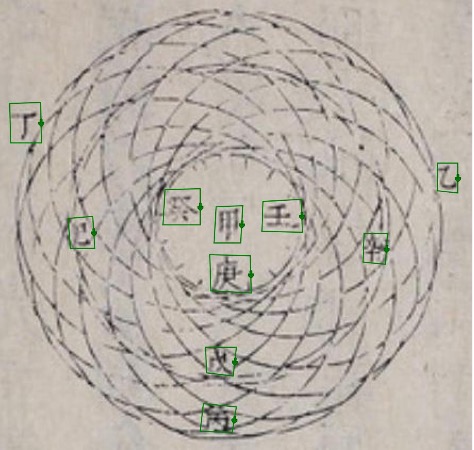} \\
        \end{tabular}
    }\end{minipage}
    }
    \caption{\textbf{Overview of our historical astronomical diagram dataset.} We contribute a large and diverse dataset of astronomical diagrams. Each text instance is annotated by a polygon consistent with reading order, indicated with directional arrow (\color{darkgreen}\faArrowRight\color{black}) symbols (\cref{subfig:a}). Our diagrams have a highly variable number of annotated text regions (\cref{subfig:b}). Our Latin diagrams' annotations are provided with text region classes (\cref{subfig:c} and~\cref{subfig:d}). The dataset includes diagrams from many traditions (\cref{subfig:e}), but diversity is very large even within a single tradition (\cref{subfig:f}).}
    \label{fig:teaser}
    \vspace{-1.5em}
\end{figure*}
Advancements in document analysis have led the research community beyond text line detection in simple documents, toward text detection in more complex scenarios, such as tables (e.g.~\cite{paliwal2019tablenet}), manga strips (e.g.~\cite{sachdeva2024manga}), maps (e.g.~\cite{li2024icdar}), and challenging historical documents. Surprisingly, despite their crucial role as reasoning tools, mathematical diagrams have received little attention. Recently, Kalleli et al.~\cite{kalleli2024historical} introduced a benchmark of 303 historical astronomical diagrams for the vectorization of lines, circles, and circular arcs, but ignored textual content. 

Historical astronomical diagrams provide an ideal testbed due to their challenging diversity. Astral sciences have been cultivated since antiquity across diverse historical contexts, each fostering unique observational practices, mathematical styles, and physical rationalizations. This diversity is reflected in the varied practices of diagrammatization and the multiple forms of mathematical, mechanical, physical, and astrological reasoning in which astronomical diagrams are involved~\cite{crowther2013training,jardine2010critical,saito2012diagrams}. As a result, the text in these diagrams exhibits remarkable diversity in language, orientation, color, function, and format, ranging from isolated symbols and single letters to full words and small paragraphs.
\\ \indent
Beyond their analytical complexity, astronomical diagrams have significant historical value. Their knowledge circulated through complex, long-range patterns, spanning centuries and connecting knowledge production across the Afro-Eurasian world. In support of these complex circulation patterns, tens of thousands of Latin, Greek, Byzantine, Arabic, Persian, Sanskrit, and Chinese manuscripts and early prints contain diagrams. They fulfilled practical, religious, and political roles, which makes them culturally significant. Furthermore, many concepts originating in the astral sciences later created other fields such as trigonometry, cartography, and meteorology. Developing tools to explore the material and epistemological dimensions of diagrams, their mutual relations, and their interaction with texts is thus essential. Text elements, along with geometric ones, are fundamental to these diagrams and critical to their interpretation~\cite{jardine2013observing,smets2009words}.
\\ \indent
We present a dataset of 948 historical astronomical diagrams annotated with 10,940 text regions. In addition, the 2,293 text regions from the 185 diagrams from the Latin tradition are annotated as single Latin characters with their transcription (1470 instances), astronomical symbols (144 instances), words (358 instances), or longer text (321 instances). We evaluate several baselines on our dataset: two recent scene text methods TESTR~\cite{zhang2022text} and DeepSolo++~\cite{ye2023deepsolo++}, and Poly-DETR a simple yet effective approach we propose to detect oriented polygonal text regions. More precisely, we modified the DINO-DETR~\cite{dinoDETR} architecture and training to predict oriented polygons, and designed a synthetic data generation pipeline to pretrain it on synthetic diagrams. We demonstrate that both elements are critical to obtain good results on our dataset. We validate this approach on standard benchmarks for text line detection in historical manuscripts and evaluate it on our dataset.
Note that our method improves the state-of-the-art performance on two text line detection benchmarks, cBAD2019~\cite{diem2019cbad} and MTHv2~\cite{wehong2020mthv2}. 
\\[5pt]
\textbf{Contributions.} In summary, our contributions are:
\raggedbottom
\begin{itemize}[leftmargin=*, topsep=0pt]
\item \textbf{A new open-access multi-linguistic dataset} of 948 historical astronomical diagrams with 10,940 annotated text regions. This dataset spans seven traditions and 10 centuries, offering high diversity for text detection in non-standard, geometrically complex layouts.
\item \textbf{Baseline evaluations} with quantitative and qualitative analysis, including Poly-DETR baseline that 
naturally handles arbitrary orientations and reading orders, \SB{not requiring large-scale pretraining on real data.}
\end{itemize}
\section{Related Work}
\label{sec:related}
\subsection{Datasets for text region detection in historical documents}
\begin{table*}[t!]
\caption{\textbf{Historical text detection datasets.} This table outlines popular historical text detection datasets which are the most closely related to ours. \textcolor{red}{\faTimes} denotes the lack of predefined splits or lack of annotations for portions of the dataset (e.g. test sets in competitions), which prevent or limit comparisons.}
\centering
\resizebox{\textwidth}{!}{
    \begin{tabular}{@{}lccccccc@{}}
    \toprule
    Dataset & Content & Access & Labels & Period & \# images & Ann. format & Script(s) \\
    \midrule
    cBAD2019~\cite{diem2019cbad} & \color{blue} \faFile*[regular]\color{black} & \color{darkgreen} \faDownload &\textcolor{darkgreen}{\faCheck} & 12$^{\text{th}}$--20$^{\text{th}}$ & 3,021 & baseline points & LAT \\
    HDRC~\cite{simistira2019hdrc}& \color{blue}\faFile*[regular] & \color{darkgreen} \faDownload & \textcolor{red}{\faTimes} & unknown & 12,850 & boxes & CHI \\
    MTHv2~\cite{wehong2020mthv2} & \color{blue}\faFile*[regular] & \color{darkgreen} \faDownload  & \textcolor{darkgreen}{\faCheck} &    unknown & 2,200 & boxes  & CHI \\
    M5HisDoc~\cite{shi2023m5hisdoc} & \color{blue}\faFile*[regular] & {\color{red} \faLock}&\textcolor{red}{\faTimes} & unknown & 4k  & boxes & CHI\\
    READ2016~\cite{toselli2018read} & \color{blue}\faFile*[regular] & \color{darkgreen} \faDownload & \textcolor{darkgreen}{\faCheck} &  14$^{\text{th}}$--19$^{\text{th}}$ & 30k & ord. polygons & LAT (DE) \\
    Bentham~\cite{sanchez2016bentham} & \color{blue}\faFile*[regular] &\color{darkgreen} \faDownload & \textcolor{darkgreen}{\faCheck} & 18$^{\text{th}}$--19$^{\text{th}}$ & 443 & ord. polygons & LAT\&GRE \\
    Saint Gall~\cite{fisher2011saintgall} & \color{blue}\faFile*[regular] & \color{darkgreen} \faDownload & \textcolor{darkgreen}{\faCheck} & $9^\text{th}$ & 60 & baseline points & LAT \\
    Balsac~\cite{vezina2020balsac}  & \color{magenta}\faObjectGroup & \color{red} \faLock & \textcolor{darkgreen}{\faCheck} & 17$^{\text{th}}$--20$^{\text{th}}$ & 913 & ord. polygons & LAT (FR) \\
    BNPP~\cite{bnpp,boillet2022robust}  & \color{cyan}\faInbox & \color{red} \faLock & \textcolor{darkgreen}{\faCheck} & 19$^{\text{th}}$--20$^{\text{th}}$ & 12 & ord. polygons & LAT (FR) \\
    HOME~\cite{boros2020home}  & \color{blue}\faFile*[regular] & \color{darkgreen} \faDownload & \textcolor{red}{\faTimes} & 12$^{\text{th}}$--15$^{\text{th}}$ & 43k & ord. polygons & LAT \\
    Horae~\cite{boillet2019horae}  & \color{blue}\faFile*[regular] & \color{darkgreen} \faDownload & \textcolor{darkgreen}{\faCheck} & 5$^{\text{th}}$--15$^{\text{th}}$  & 572 & ord. polygons & LAT \\
    ScribbleLens~\cite{dolfing2020scribble} & \color{gray}\faShip & \color{darkgreen} \faDownload & \textcolor{red}{\faTimes} & 16$^{\text{th}}$--18$^{\text{th}}$ & 1k & boxes & LAT (NL) \\
    Rumsey~\cite{lin2024rumsey} & \color{orange}\faMap[regular] & \color{darkgreen} \faDownload & \textcolor{darkgreen}{\faCheck}         & 16$^{\text{th}}$--21$^{\text{th}}$ & 940 & ord. polygons  & LAT (EN) \\
    FLR~\cite{chazalon2024ign} & \color{orange}\faMap[regular] & \color{darkgreen} \faDownload & \textcolor{darkgreen}{\faCheck} & 19$^{\text{th}}$ & 145 & ord. polygons  & LAT (FR) \\
    TMS~\cite{lin2025tw} & \color{orange}\faMap[regular] & \color{darkgreen} \faDownload & \textcolor{darkgreen}{\faCheck} & 20$^{\text{th}}$ & 1,644 & ord. polygons & CHI \\
    \midrule
    \textbf{Ours} & \color{darkgreen}\astroicon{} & \color{darkgreen} \faDownload & \textcolor{darkgreen}{\faCheck} & 8$^{\text{th}}$--18$^{\text{th}}$& 948 & ord. polygons & \color{LimeGreen} \faGlobe \\
    \bottomrule
    \end{tabular}
    }
    \vfill
    \resizebox{\textwidth}{!}{
    \begin{tabular}{rl@{\;\;\;\;}rl@{\;\;\;\;}rl@{\;\;\;\;}rl}
        \color{blue}\faFile*[regular] & historical documents  
        & \color{orange}\faMap[regular] & historical maps 
        & \textcolor{darkgreen}{\astroicon{}} & historical astronomical diagrams 
        & \color{darkgreen}\faDownload & open-access \\
        \textcolor{darkgreen}{\faCheck} & labels available 
        & \textcolor{red}{\faTimes} & no/partial labels 
        & \color{LimeGreen} \faGlobe & CHI, AR, LAT, GRE, HEB, PER, SAN  & \color{red} \faLock & no access
        \\
        \textcolor{gray}{\faShip} & ship journals  & 
        \textcolor{cyan}{\faInbox} & bank archives & \textcolor{magenta}{\faObjectGroup} & death/birth/marriage records \\
        \end{tabular}}
    \label{tab:dataset}
    \vspace{-1em}
\end{table*}
Text detection aims at localizing text instances within images using representations such as bounding boxes, polygons, or Bézier curves. This task is traditionally split between \textit{scene text detection}, which handles arbitrary orientations and complex backgrounds in natural images, and \textit{document text detection}, which focuses on scanned pages with horizontally aligned texts. Historical documents bridge these areas with non-standard layouts and pose unique challenges with material degradation and densely-packed handwritten instances. \cref{tab:dataset} summarizes key historical datasets providing text-level annotations across domains such as historical manuscripts~\cite{cbad,diem2019cbad,li2024icdar,simistira2019hdrc,wehong2020mthv2,jian2023hisdoc}, historical maps~\cite{lin2024rumsey,chazalon2024ign,lin2025tw}, bank archives~\cite{bnpp,boillet2022robust}, ship journals~\cite{dolfing2020scribble}, and personal records~\cite{vezina2020balsac}. Although these datasets cover diverse scripts and layouts, they largely focus on text in linear, cartographic, or tabular contexts. To our knowledge, our dataset is the first to specifically address historical astronomical diagrams, which complements the linear flow of journals or the geographic spread of maps. Furthermore, our dataset offers extensive temporal coverage and high linguistic diversity, spanning ten centuries and six different scripts.

\subsection{Methods for text detection in historical documents}
The methods for text detection in historical documents vary primarily by their output representation, including baselines~\cite{oliveira2018dhsegment,gruning2019two}, segmentation masks~\cite{monnier2020docextractor,sharan2021palmira,fizaine2024historical,gao2024kesar}, and bounding boxes~\cite{wehong2020mthv2,jian2023hisdoc}. Although convolutional architectures for detection~\cite{moysset2016learning} and segmentation~\cite{fink2018baseline,islam2023line} saw early success, recent years have seen a transition towards Transformer-based architectures inspired by DETR~\cite{carion2020end}. Examples include CURT~\cite{kiessling2022curt} for baseline coordinates and LineTR~\cite{agrawal2025linetr} for parametric scribbles.
\looseness=-1

Standard document methods often assume a predominantly horizontal alignment, which is insufficient for the complex layouts of scientific diagrams. In contrast, scene text detection methods localize arbitrarily-shaped text in natural images. Models such as TESTR~\cite{zhang2022text}, DeepSolo~\cite{ye2023deepsolo++} or LRANet++~\cite{su2025lranet++} utilize polygons, Bézier curves or center-line points within transformer-based frameworks to capture curvature and rotation. 
These approaches are also relevant for historical maps, where research has adapted scene text models~\cite{long2021scene,li2023mask} and synthetic data~\cite{li2021synthetic,lin2024hyper} to handle high-density, multi-oriented text in cartographic contexts~\cite{chiang2016unlocking,kim2023mapkurator}.

Following this trend, we consider as baseline TESTR~\cite{zhang2022text} and DeepSolo++~\cite{ye2023deepsolo++}, which pretraining on large-scale scene-text datasets is critical, as well as a simple baseline, Poly-DETR, which we build by adapting DINO-DETR~\cite{dinoDETR} to regress ordered polygon vertices and pretrain on a simple synthetic diagram-like dataset. 
\section{Historical astronomical diagrams dataset}
\label{sec:dataset}
\begin{figure*}[t]
    \centering
    \vspace{-1em}
    \resizebox{0.99\linewidth}{!}{
    \centering
    \begin{minipage}{\textwidth} 
   \subfloat[Aspect ratio\label{subfig:statb}]{
        \includegraphics[width=0.33\textwidth]{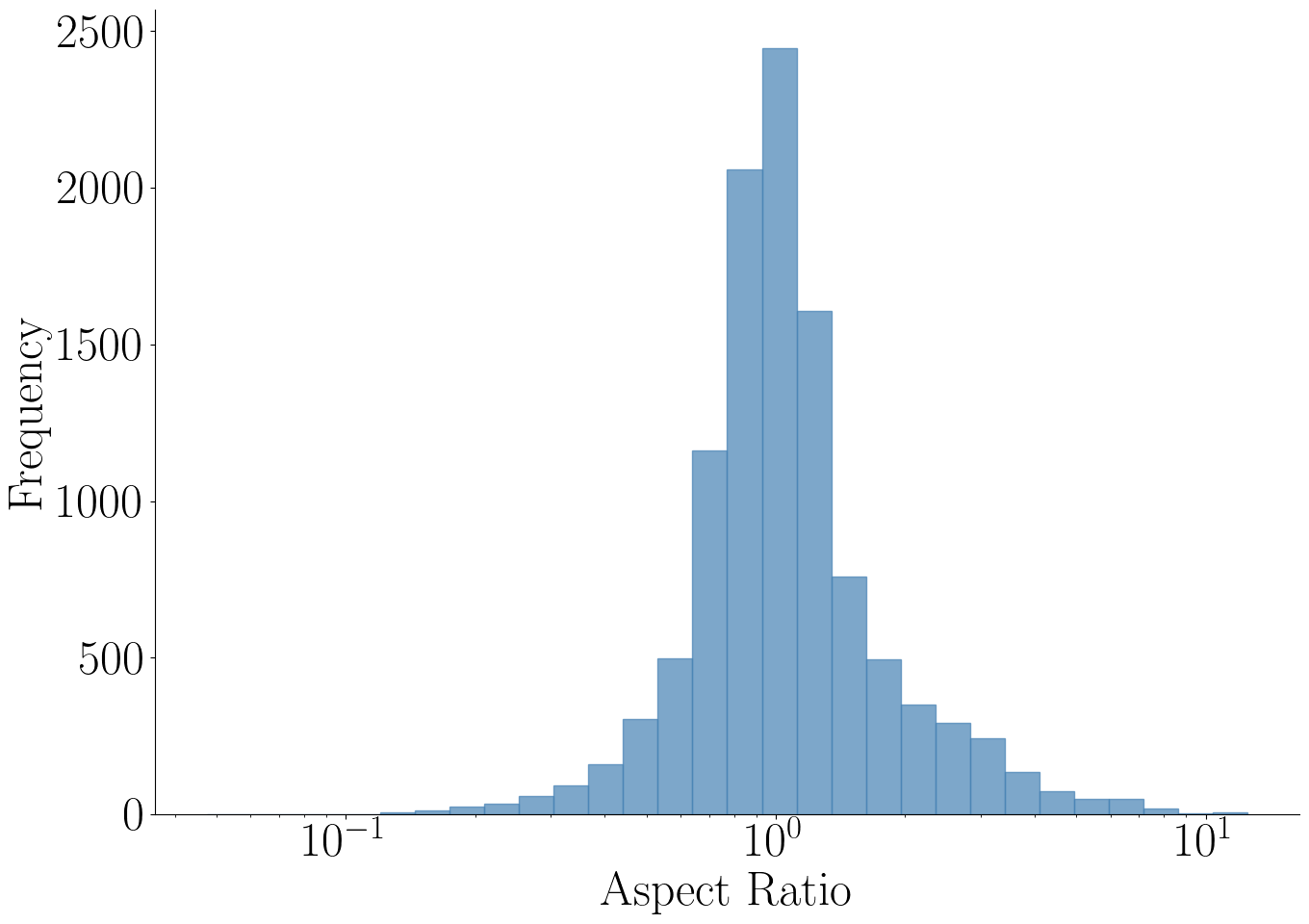} 
    } 
    \subfloat[Area w.r.t. image size\label{subfig:statd}]{
     \includegraphics[width=0.33\textwidth]{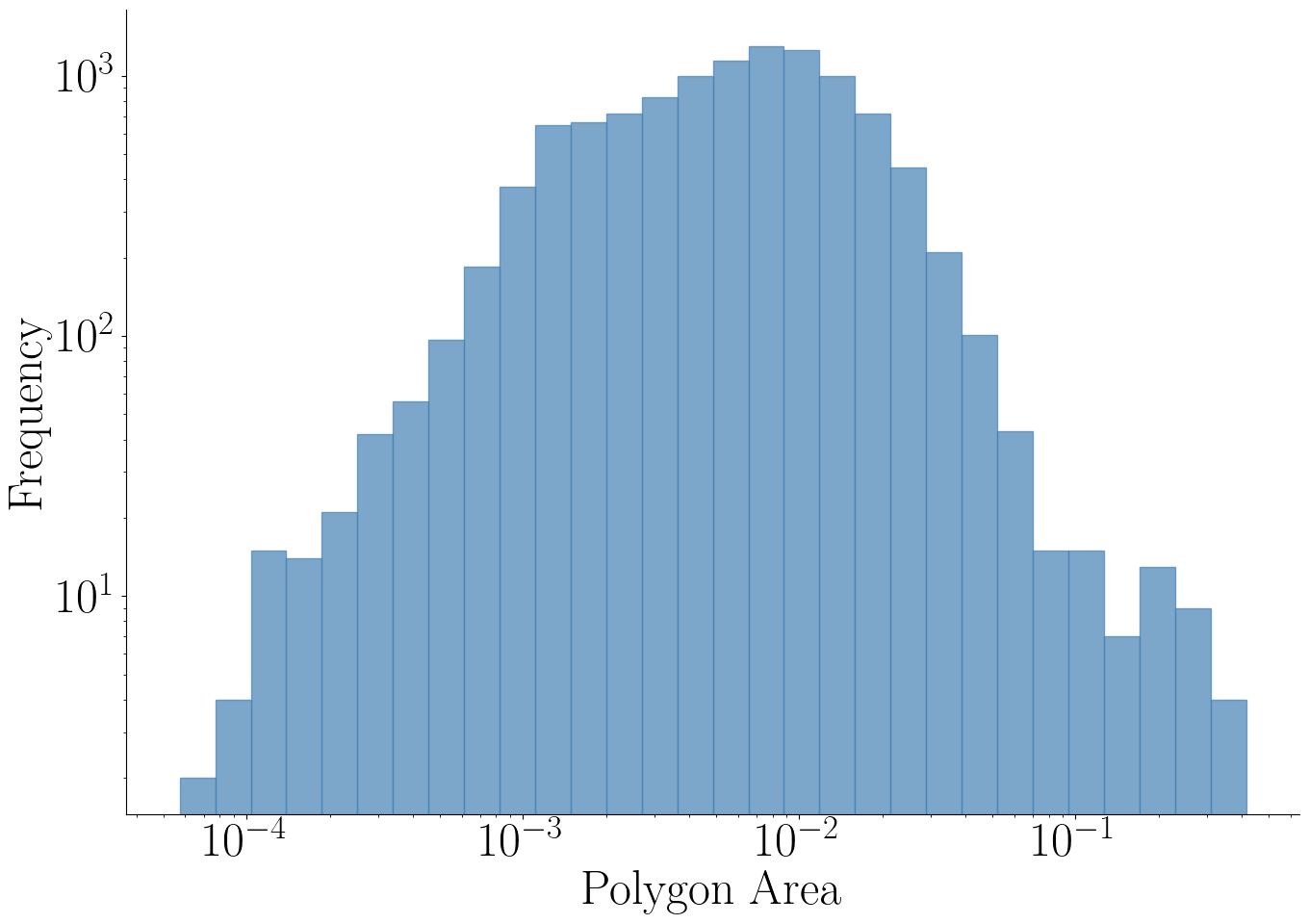}}
    \subfloat[Spatial distribution\label{subfig:state}]{
     \includegraphics[width=0.33\textwidth]{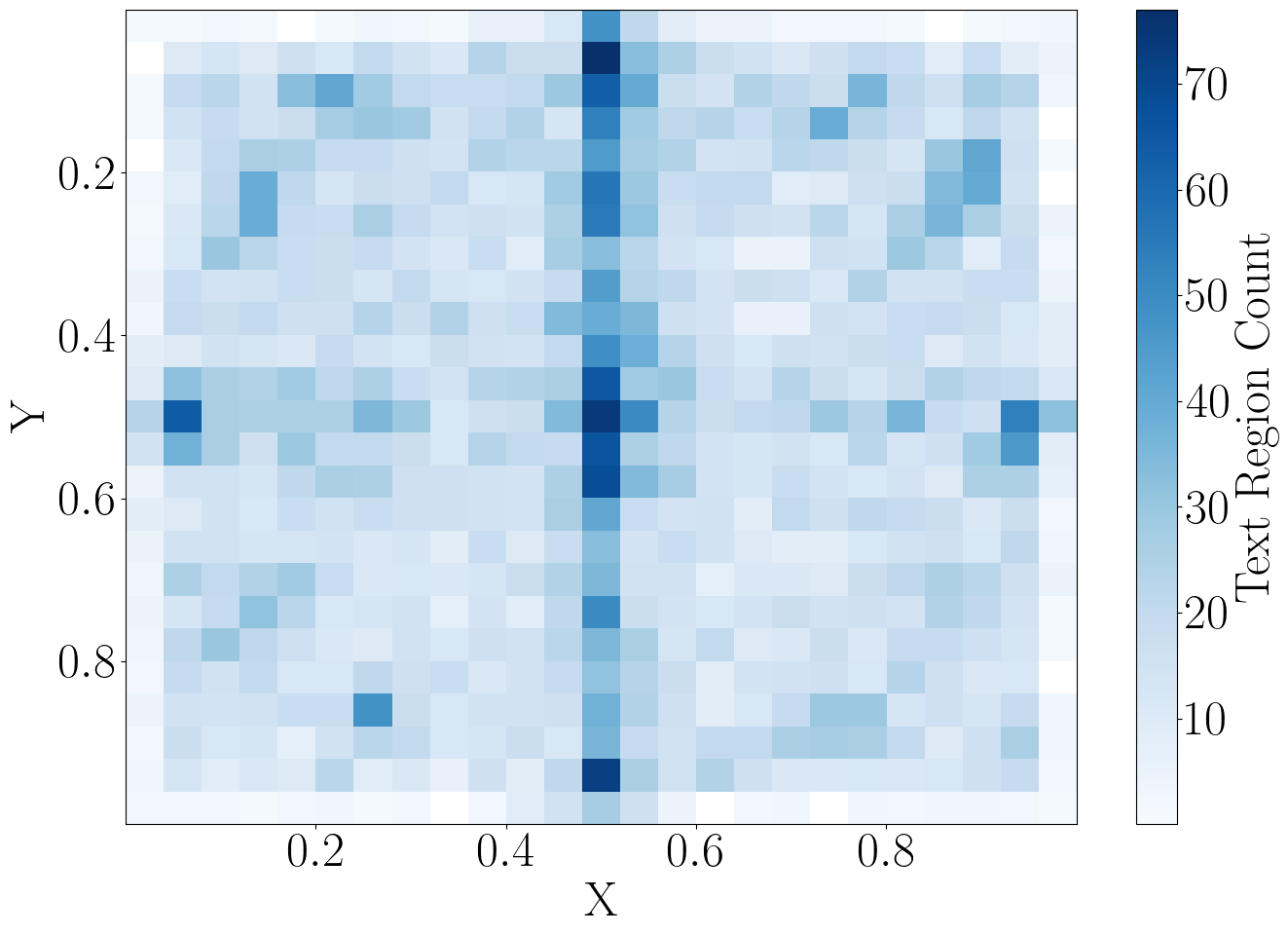}}
     \end{minipage}
}
    \caption{\textbf{Text regions distributions in our dataset.} The aspect ratio, size relative to the full image size, and position in the image of our text region have a high diversity.}
    \label{fig:stats}
    \vspace{-1em}
\end{figure*}
\subsection{Data and annotations}
We present qualitative examples of diagrams and annotations in~\cref{fig:teaser}, and some statistics from our dataset in~\cref{fig:teaser,fig:stats}. 
\\[5pt]
\textbf{Positioning.}
Historical documents containing text instances are diverse and can be broadly categorized into: (i) \textit{textual documents}, including handwritten manuscripts, letters, archival records, and printed materials such as newspapers and administrative records; (ii) \textit{cartographic documents}, such as maps and city plans, where text often follows arbitrary orientations; and (iii) \textit{technical and scientific illustrations}, including diagrams, technical drawings, and charts, where text is typically overlaid on images, often appearing in arbitrary shapes and orientations. Although there is an increasing number of datasets for textual and cartographic documents (such as the ones listed in~\cref{tab:dataset}), to our knowledge, there is no dataset for text detection in historical diagrams. 
To remedy this, we present an open-access dataset of 948 historical astronomical diagrams, exhaustively annotated with 10,940 text regions.
\\[5pt]
\textbf{Diagrams selection.}
Together with expert historians, we selected a diverse set of astronomical diagrams ranging from the 8$^{\text{th}}$ to the 18$^{\text{th}}$ century and spanning seven major traditions: Persian and Arabic (PER, AR, 115), Chinese (CHI, 332), Byzantine (GRE, 233), Latin (LAT, 185), Hebrew (HEB, 48), and Sanskrit (SAN, 35). 
This outstanding temporal and cultural breadth provides rich variation across multiple dimensions: content representation, script systems, artistic styles, material substrates, digitization quality, and preservation conditions, as visualized in~\cref{subfig:e,subfig:f}.
\\[5pt]
\textbf{Ordered polygon annotations for multi-lingual class-agnostic diagram text localization.}
Historians annotated diagrams with ordered polygons that bound text and encode reading sequences via vertex order. Each polygon concatenates a bottom and a top line with the same number of points: the first half follows the reading order along the bottom, while the second half returns in the opposite direction along the top.
To unify the annotations and simplify the processing, we resampled the lines into $K=32$ points, equally and regularly sampled along the baseline and the top line, verifying that polygon geometry remained consistent. This explicit ordering sets our dataset apart from other historical manuscript text line segmentation datasets, where the annotation of the reading order is mostly absent. However, it is particularly important for diagrams, where text could be written in any direction, as can be seen in~\cref{subfig:a}. In total, 10,940 text regions were annotated. 
Note that the diagram images might include text that is not related to the diagram, which we do not annotate since our goal is to extract only the information directly relevant to the diagram. An alternative could have been to annotate the diagram area segmentation. However, this can be challenging in practical cases, such as the leftmost diagram of~\cref{subfig:e}, where the diagram and the external text are intertwined, and the separation between both types of text requires advanced diagram structure understanding, which we do not believe should be separated from diagram text identification.
{We split our dataset into 760 diagrams for training, 96 diagrams for validation, and 92 diagrams for testing. We use it to evaluate text localization and reading-order prediction in a class-agnostic way, i.e., independently of the text content.}
\\[5pt]
\noindent \textbf{Content annotations for class-aware text localization in Latin diagrams.}
A more advanced task would be to read the diagram text. Since multi-lingual, multi-script transcription seems out of reach, we focused on a simpler text classification task on our 185 Latin diagrams. We first differentiated single-character (1614 instances), word (358 instances), and longer sentence (321 instances) annotations. Single characters were also annotated, resulting in 49 class labels: 23 lowercase characters, 14 uppercase characters, 10 digits, an astronomical symbol class (144 instances), and an additional class for unreadable characters. Because many characters appear very few times in our dataset, we grouped characters that appeared less than 30 times in a single class. This resulted in 20 classes, including 16 character classes, the symbol, word and longer sentence classes, and a specific class for unreadable characters (\cref{subfig:d}). We use these annotations to evaluate class-aware text localization in Latin diagrams. {We split this Latin tradition dataset with class labels into 100 diagrams for training, 40 diagrams for validation, and 45 diagrams for testing.}
\\[5pt]
\textbf{Statistics.} {As shown in~\cref{subfig:b}, the number of text regions in the diagrams varies a lot from 1 to 207 polygons per diagram with a median of 8. The sizes of the text regions also have high diversity, both in terms of aspect ratio (\cref{subfig:statb}) and size compared to the full size of the diagram image (\cref{subfig:statd}).~\cref{subfig:state} visualizes the spatial distribution of the coordinates of the center of the text regions relative to the size of the image. Although there is a clear bias toward the median vertical line, which is expected, there is still a lot of diversity.}

\subsection{Evaluation metrics}
\textbf{F1 score and Average Precision.}
We use the F1 score and Average Precision to evaluate class-agnostic text localization over our complete dataset. We assume that algorithms predict polygons associated with confidence scores, and  setting a threshold on the confidence score thus produces a set of predicted polygons. Given a set of predicted polygons and a set of ground truth text polygons for an image, we first use Intersection over Union (IoU) to define potentially valid associations, setting a threshold of 0.5. We then rely on greedy matching over confidence scores 
to associate each ground truth polygon with at most one predicted polygon, maximizing the sum of the IoU distance between the associated polygons. 
From these, we compute Precision (P), Recall (R), and the F1 score (F1) using the standard definitions. We then integrate the precision-recall curve to obtain average precision (AP50), and report the best F1 score by selecting the optimal threshold on the validation set and computing the associated score on the test set.

In the class-aware case, we compute average precision independently for each class and then average the results to obtain mean average precision (mAP50). Similarly, for each threshold, we compute the average of the per-class F1 score (mF1), select the best threshold on the validation set, and report the associated performance on the test set.
\\[5pt]
\textbf{Metrics with Reading Order.}
Although reading order is crucial for many applications, existing datasets rarely annotate it and no standard metric accounts for it. We address this by defining {F1-O} and AP50-O, simple variants of F1 and average precision scores that only count as positive detections whose reading order is correct. To compute it, we first obtain a set of candidate true positive matches by matching predictions with ground truth polygons in the same way as for the standard score.  
We then filter those for which the reading order is consistent between prediction and ground truth polygons: we consider the nearest neighbors among the 4 ground truth polygon corners of the 4 prediction polygon corners and only keep the match if they are correct. Finally, using only the filtered matches as positive, we compute F1-O and AP50-O in a way similar to the standard metrics defined above. 

\subsection{Annotation verification and consistency}
\begin{wrapfigure}{r}{0.5\textwidth}
\vspace{-3em}
\begin{center}
    \includegraphics[width=0.48\textwidth]{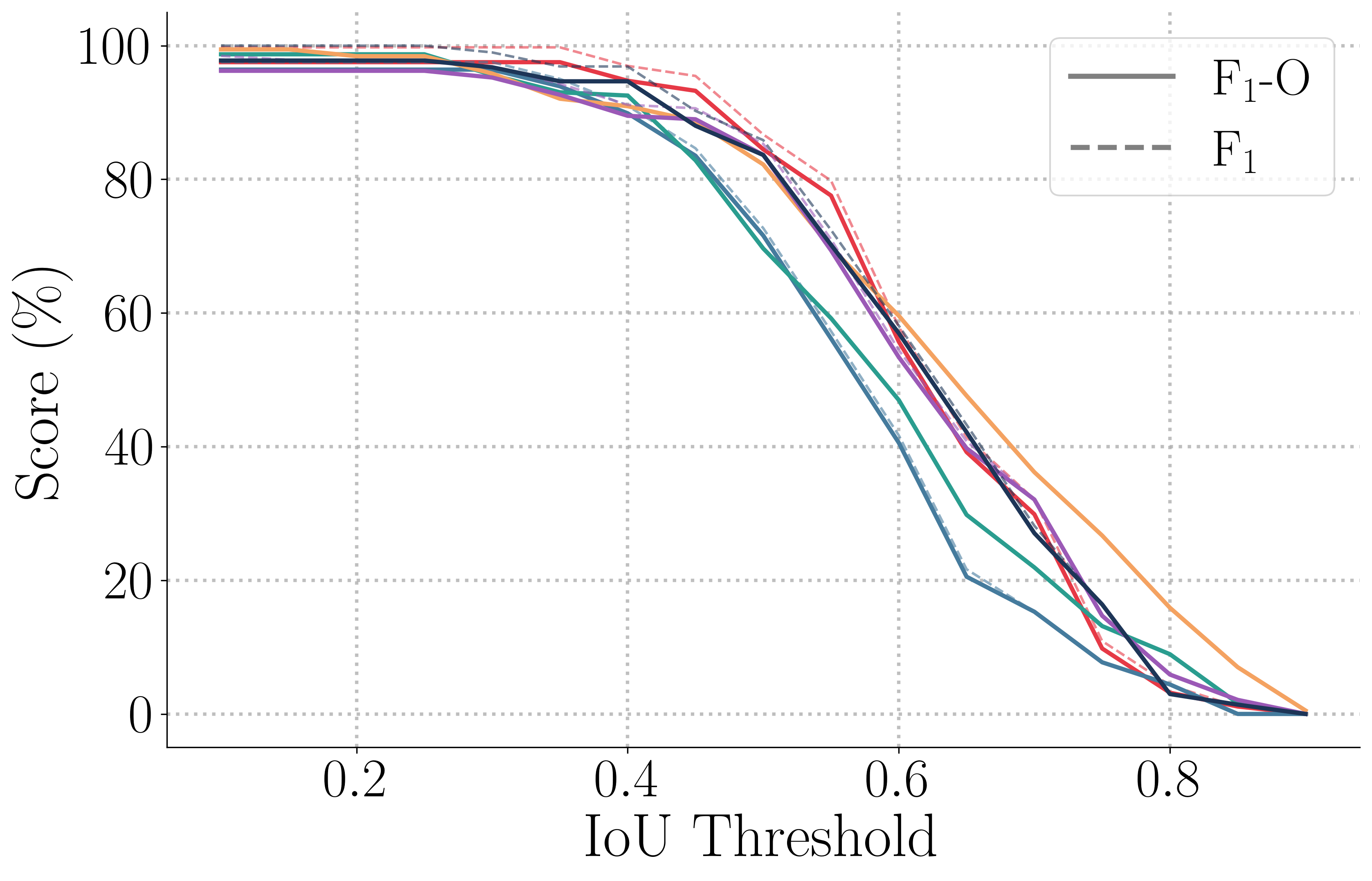}
  \end{center}
    \caption{\textbf{Annotation comparison.} 6 annotations (colored) on a subset of diagrams are compared with the corresponding ground truth (expert) annotations.}
    \label{fig:annotation}
\vspace{-2em}
\end{wrapfigure}
Region annotations were done in multiple steps: a first set of annotators, including experts in each of the traditions, made a first version of the annotations, from which a clear visualization of the resulting annotation, similar to the one used in this paper, was generated; then, a second set of annotators validated those results and either fixed small errors or simply discarded diagrams with wrong annotations.
To minimize errors in the final dataset, the last step was repeated twice.

To assess the consistency of annotators -- and thus the quality of our annotations -- and better understand the effect of thresholding positive with a mIoU of 0.5, we had 6 annotators re-annotate a set of 20 random diagrams from the Latin tradition. The F1 and F1-O scores are shown as a function of the IoU threshold used to define positive matches in \cref{fig:annotation}. Given that overlaps are rare between text regions, scores close to 100\% below a threshold of 0.3 show that annotators agree on the definition of a text region, and the fact that F1 and F1-O scores are almost overlapping shows that they almost always agree on orientation. The quick decrease in the F1 score after a threshold of 0.5 underlines the difficulty in defining and annotating historical handwritten text regions beyond regular lines and bounding boxes. Indeed, the annotators did not consistently consider large ascenders, descenders, or decorators. We selected the standard threshold of 0.5 to compute our metrics, for which 4 out of our 6 test annotators obtained an F1-O score above 80\%, but it is actually much more challenging than for standard detection tasks.
\section{Baselines}
\label{sec:method}
\begin{figure*}[t!]
    \centering
    \includegraphics[width=\linewidth]{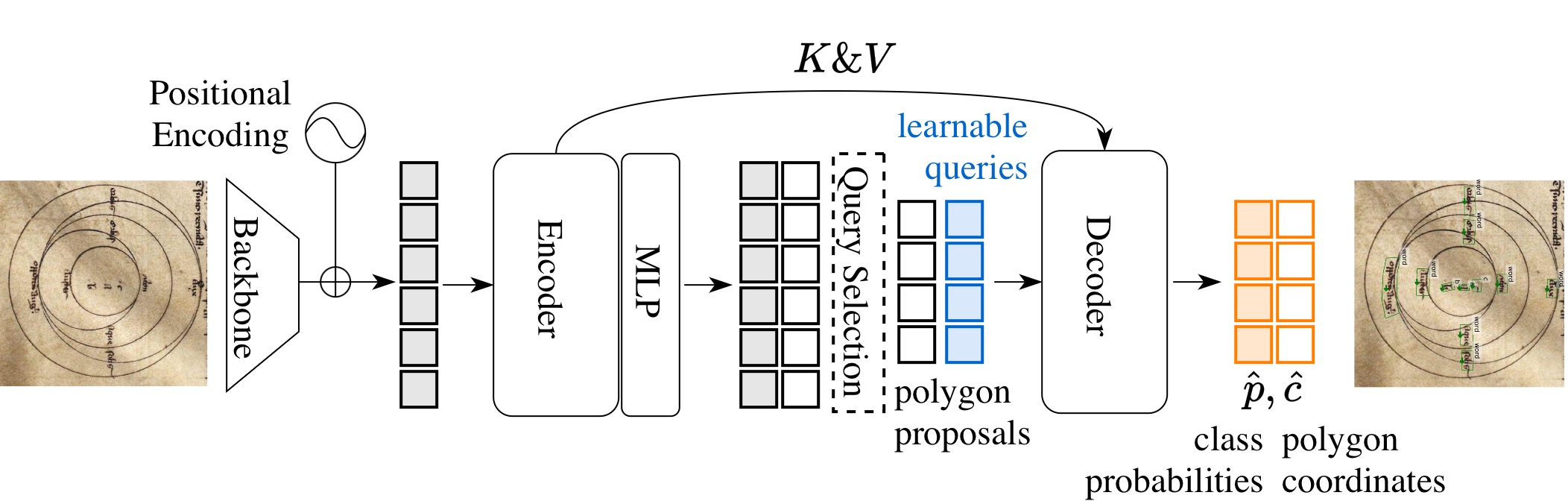}
      \caption{{\bf Poly-DETR.} Given image features from a backbone, a transformer encoder predicts initial anchors and tokens, that are used as queries by a transformer decoder to predict, for each query token $q$, ordered polygon coordinates $\hat{\mathbf{c}}_q$ and a probability $\hat{p}_q$.}
    \label{fig:arch}
    \vspace{-1.5em}
\end{figure*}
In the following section, we present the methods we selected as baselines, first the TESTR~\cite{zhang2022text} and DeepSolo++~\cite{ye2023deepsolo++} scene text detection approaches (\cref{subsec:baselines}), then our proposed simple extension of DINO-DETR, Poly-DETR, with its training and implementation details (\cref{subsec:arch}).
\subsection{TESTR and DeepSolo++}
\label{subsec:baselines}
{We evaluate two standard text detection baselines that utilize polygon-based representations on our dataset: (i) TESTR~\cite{zhang2022text}, a widely adopted transformer-based framework that employs dual decoders to regress polygon control points, offering a robust baseline for multi-oriented text detection, and; (ii) 
DeepSolo++~\cite{ye2023deepsolo++} 
for multi-lingual text detection with a transformer-based architecture. Both models were selected due to their native encoding of the reading order of text instances, a feature critical for the non-linear multi-linguistic texts in our dataset.
We also considered LRANet++~\cite{su2025lranet++} which excels at scene text detection, but its requirement for a consistent canonical starting vertex led to unstable training on our dataset. 
To assess the importance and impact of training data, we evaluate both TESTR~\cite{zhang2022text} and DeepSolo++~\cite{ye2023deepsolo++} under three regimes: trained only on large-scale scene-text data, trained only on our dataset, and pretrained on scene-text before finetuning on our dataset.}

\subsection{Poly-DETR}
\label{subsec:arch}

To provide a strong simple baseline, which does not require pretraining on large-scale scene text, we present a simple adaptation DINO-DETR~\cite{dinoDETR}, a standard transformer-based encoder-decoder architecture designed for object detection. 
\\[5pt]
\textbf{Architecture.} 
To detect text regions of arbitrary shapes, we modify the bounding box prediction layers of DINO-DETR to instead output polygon coordinates (\cref{fig:arch}). 
Formally, our transformer decoder processes $Q$ queries and outputs $Q$ ordered polygon vertices $\hat{\mathbf{c}}_q\in \mathbb{R}^{2K}$, and probabilities $\hat{{p}}_q\in [0,1]$, where $K$ is the (fixed) number of polygon vertices and $q\in\{1, ... , Q\}$ is the query index.
For class-agnostic text region detection, our model outputs a single probability, and for class-aware detection, it outputs a probability for each class independently. 
\\[5pt]
\textbf{Loss.} 
To transition from bounding boxes to polygons, we first omit the Intersection over Union (IoU) term from both the loss and the Hungarian matching, as computing a fully differentiable IoU for non-convex polygons is computationally challenging. Second, we adapt the denoising strategy used during training to polygons and noise each polygon vertex randomly and independently.

More formally, in the class-agnostic scenario, our loss compares $Q$ predictions $\{(\hat{\mathbf{c}}_q, \hat{p}_q)\}_{q\in\{1,...Q\}}$ to a set of $N$ ground truth polygons of coordinates ${\mathbf{c}_n}\in \mathbb{R}^{2K}$ for each image, where $n\in\{1,...N\}$ is the index of the ground truth polygon and $K$ is the number of polygon vertices. We assume $Q>N$ and define $p_n=1$ if $n\{1,...N\}$ and $p_n=0$ if $n\in\{N+1,...,Q\}$.
The network is trained by first finding an optimal permutation $\hat{\sigma}$ of the Q queries using the Hungarian algorithm to minimize the bipartite matching cost,
%
%
where the pairwise cost is defined as:
\begin{equation}
\mathcal{L}_{\text{cost}}(p_n, \mathbf{c}_n, \hat{p}_q, \hat{\mathbf{c}}_q) = \lambda_{\text{cls}} \mathcal{L}_{\text{cls}}(p_n, \hat{p}_q) + \mathbf{1}_{{n \leq N}} \lambda_{\text{poly}} \mathcal{L}_{\text{poly}}(\mathbf{c}_n, \hat{\mathbf{c}}_q),
\end{equation}
were, $\mathcal{L}_{\text{cls}}$ is the focal loss~\cite{lin2017focal}, $\mathcal{L}_{\text{poly}}$ is the $L^1$ distance, and $\lambda_{\text{cls}}$ and $\lambda_{\text{poly}}$ are two scalar hyperparameters. Once matched, the final training loss $\mathcal{L}$ follows the same functional form as $\mathcal{L}_{\text{cost}}$, using the optimal assignment $\hat{\sigma}$ and the updated hyperparameters $\lambda'_{\text{cls}}$ and $\lambda'_{\text{poly}}$. For class-aware detection, the loss is applied to each class independently. 
\\[5pt]
\textbf{Training and implementation details.} 
To pretrain our architecture, we generate synthetic diagrams by adapting the code of~\cite{kalleli2024historical}. We extend it to incorporate texts in Latin, Greek, Arabic, Chinese and Hebrew scripts
with orientation, slant, and other stylistic attributes and to provide ordered polygons for each text instance. 
\SB{We generate sets of 2,500 synthetic samples, each reused 10 times with varied augmentations before regeneration. We used as script distribution: Arabic (40\%), Latin (30\%), Chinese (15\%), Greek (10\%), Hebrew (5\%).}
For class-aware text region localization in Latin script, we created a synthetic dataset with text regions labeled with 19 classes that we define in our dataset, excluding the non-digital category ``symbol''. 

We follow~\cite{dinoDETR} for architecture hyper-parameters. 
We use the ADAM optimizer with a batch size of 4, $\beta_1=0.9$, $\beta_2=0.999$, and a weight decay of $10^{-4}$. For class-agnostic text region detection, we first pretrain our model on our synthetic dataset for 750k iterations with a 
learning rate of $10^{-4}$ and 250k iterations with a learning rate of $10^{-5}$. 
We then finetune the model on our dataset for 16k iterations with a learning rate of $10^{-4}$ and 16k iterations with a learning rate of $10^{-5}$. 
For class-aware text region detection, we pretrain our model on synthetic data with the same recipe as class-agnostic text region detection. Then, we finetuned our model on the real dataset for 1,5k iterations with a learning rate of $10^{-4}$ and 1,5k iterations with a learning rate of $10^{-5}$. For finetuning, we reinitialized the classification head for 20 classes to include the additional ``symbol'' class.
\\[5pt]
\textbf{Adaptation for text line benchmarks.}
\label{subsec:beyond}
To validate the efficiency of our Poly-DETR baseline, we evaluate it on standard historical text line benchmarks~\cite{diem2019cbad,wehong2020mthv2,simistira2019hdrc}. We standardize all annotations to a $K=10$ vertex polygon format. 
By achieving competitive results on structured documents, our goal is to demonstrate that Poly-DETR is a strong baseline and its performance on astronomical diagrams reflects the inherent difficulty of the dataset rather than its own limitations.

For the documents we pretrain the model for 350k iterations on the synthetic data of~\cite{monnier2020docextractor} and finetune it with a smaller batch size of 2 to fit large documents in memory. We kept a learning rate of $10^{-4}$ when finetuning on MTHv2~\cite{wehong2020mthv2} and HDRC~\cite{simistira2019hdrc}, which are qualitatively quite different from our pretraining data, and decreased it to $10^{-5}$ in the middle of training for cBAD~\cite{diem2019cbad}. We also adapted the number of finetuning iterations for each dataset, 250k for cBAD~\cite{diem2019cbad}, 175k for MTHv2~\cite{wehong2020mthv2} and 1M for HDRC~\cite{simistira2019hdrc}.

\section{Results}
\label{sec:results}
In this section, we validate Poly-DETR on standard text line localization benchmarks (\cref{sec:res_other}), and report baseline results on our astronomical diagram dataset in class-agnostic (\cref{sec:res_diag}) and class-aware scenarios (\cref{sec:res_cls_diag}).

\subsection{Poly-DETR approach validation for text line localization} 
\label{sec:res_other}
\begin{table}[t!]
    \centering
    \caption{\textbf{Quantitative results for text line detection} on the cBAD2019~\cite{diem2019cbad}, MTHv2~\cite{wehong2020mthv2}, and HDCR19~\cite{simistira2019hdrc} datasets. (*) HDRC lacks an official test split, making results incomparable; we report 5-fold cross-validation.}
    \label{tab:res-baseline}   
    \begin{tabular}[b]{c}
    \resizebox{0.36\textwidth}{!}{
    \begin{tabular}{lccc}
    \toprule
     Method & \multicolumn{3}{c}{cBAD2019~\cite{diem2019cbad}} \\
     \midrule
     & {P} & {R} & {F1} \\        
     DMRZ~\cite{fink2018baseline,diem2019cbad} & $92.5$ & $90.5$ & $91.5$ \\
     Planet~\cite{diem2019cbad} & $93.7$ & $92.6$ & $93.1$ \\
     docExtractor~\cite{monnier2020docextractor} & $92.0$ & $93.1$ & $92.5$ \\
    \midrule
     Poly-DETR & $\mathbf{94.2}$ & $\mathbf{93.9}$ & $\mathbf{93.9}$ \\
     \bottomrule
    \end{tabular}}
    \end{tabular}
    \begin{tabular}[b]{c}
    \resizebox{0.6\textwidth}{!}{
    \begin{tabular}{lcccccc}
      \toprule
     Method &  \multicolumn{3}{c}{MTHv2~\cite{wehong2020mthv2}} & \multicolumn{3}{c}{HDRC~\cite{simistira2019hdrc}$^*$} \\
     \midrule
     & {P} & {R} & {F1} & {P} & {R} & {F1} \\
     KESAR~\cite{gao2024kesar} & 93.4 & 93.1 & 93.2 & $-$ & $-$ & $-$ \\
     Mask R-CNN~\cite{he2017mask,hu2024seghist} & $98.2$ & $96.0$ & $97.1$ & $96.6$ & $96.2$ & $96.4$ \\
    Deformable DETR~\cite{zhu2020deformable,hu2024seghist} & $97.9$ & $94.6$ & $96.3$ & $94.4$ & $95.7$ & $94.6$ \\
    PAN~\cite{PAN,hu2024seghist} & $97.2$ & $93.1$ & $95.1$ & $95.1$ & $92.8$ & $94.0$ \\
    OBD~\cite{OBD,hu2024seghist} & $97.8$ & $97.3$ & $97.6$ & $94.6$ & $97.0$ & $95.7$ \\  
    DTDT~\cite{hu2024seghist} & $97.9$ & $97.9$ & $97.9$ & $96.9$ & $96.4$ & $96.6$ \\
    \midrule
     Poly-DETR & $\mathbf{98.4}$ & $\mathbf{98.5}$ & $\mathbf{98.4}$ & $96.7$ & $96.5$ & $96.6$ \\
     \bottomrule     
    \end{tabular}}
    \end{tabular}
    \vspace{-1.5em}
\end{table}
%
%
We evaluate Poly-DETR on three standard historical datasets: cBAD2019~\cite{diem2019cbad} (Latin), MTHv2~\cite{wehong2020mthv2} (Chinese), and ICDAR 2019 HDRC~\cite{simistira2019hdrc} (Chinese). cBAD2019 provides 3,021 European archival documents with baseline and bounding box annotations. MTHv2 contains 5,598 documents with bounding boxes. HDRC includes 11,715 documents with bounding boxes, but it lacks an official test split; therefore, we use 5-fold cross-validation for our model and compare with results reported in~\cite{dtdt}. Quantitative results in \cref{tab:res-baseline} show that our method outperforms the state-of-the-art on cBAD2019 and MTHv2, while achieving competitive results on HDRC. Qualitative analysis reveals failure modes similar to those we will see in our astronomical diagrams: errors in text grouping and confusion between text and graphical elements.
\subsection{Class-agnostic text region detection in astronomical diagrams}
\label{sec:res_diag}
\begin{table*}[t!]
    \caption{\textbf{Quantitative results for class-agnostic text detection on historical astronomical diagrams.} \textit{pt.:} pretraining, \textit{ft.:} finetuning.
    }
    \centering    
    \tabcolsep=5pt
    \small
    \begin{tabular}{@{}lcccccc@{}}
    \toprule
     Method & \textit{pt.} & \textit{ft.} &  {F1} & {F1-O} & {AP50} & {AP50-O}\\
     \midrule
     TESTR~\cite{zhang2022text} & scene text & \checkmark  & \bf 80.4 & \bf 71.1 & \bf 82.4 & \bf 68.3 \\
     ~ w/o finetuning & scene text &  & 30.6 & 9.7 & 14.8 & 0.0 \\ 
     ~ w/o pretraining & none & \checkmark& 47.4 & 37.6 & 36.2 & 24.8 \\
    Poly-DETR & synthetic & \checkmark & {79.7} & \underline{69.1} & \underline{76.6} & {61.9} \\ 
    ~ w/o finetuning & synthetic &  & 39.8 & 10.9 & 28.2 & 2.7 \\
    ~ w/o pretraining & none & \checkmark &  {73.2} & {68.6} & {70.3} & \underline{62.1} \\
     ~ w/ 4 points & synthetic & \checkmark & 72.1 & 48.9 & 67.8 & 34.3 \\
    DeepSolo++~\cite{ye2023deepsolo++} & scene text & \checkmark & \underline{80.0} & 61.7 & 75.6 & 49.4 \\
     ~ w/o finetuning & scene text &  & 24.3 & 6.4 & 13.8 & 1.3 \\
     ~ w/o pretraining & none & \checkmark & 0.0 & 0.0 & 0.0 & 0.0 \\
     \bottomrule
    \end{tabular}
    \vspace{-1.5em}
    \label{tab:res-new}
\end{table*}

For class-agnostic text region detection on our dataset, we evaluated three approaches that enable predicting polygons: TESTR~\cite{zhang2022text} and DeepSolo++~\cite{ye2023deepsolo++} scene text detection models and our new baseline Poly-DETR. Note that TESTR~\cite{zhang2022text} and DeepSolo++~\cite{ye2023deepsolo++} are pretrained on a very large database of scene text images to jointly predict text localization and transcription.
The results in~\cref{tab:res-new} suggest that TESTR and PolyDETR adapt much better than DeepSolo++ to text directionality.
Poly-DETR demonstrates superior performance, suggesting a higher level of sample efficiency. It is, however, slightly overperformed by TESTR pretrained on large-scale scene text data.  
Interestingly, using Poly-DETR with only $K=4$ vertices offers results approximate with those obtained with more vertices when one does not consider reading order, but much lower when considering reading order, highlighting the importance of computing metrics with reading order. While TESTR and PolyDETR successfully converge when trained from scratch, DeepSolo++'s Bézier control and boundary point optimization fail to converge without large scale pretraining when faced with diverse reading orders. In general, our results underscore the benefits of pretraining.
\begin{wraptable}{r}{0.4121\textwidth}
    \centering
    \vspace{-2em}
    \caption{\SB{\textbf{Per-script results} of Poly-DETR on class-agnostic text region detection.}}
    \tabcolsep=4pt
    \small
    \begin{tabular}{@{}lccc@{}}
    \toprule
         Script & Count & F1 & F1-O \\
         \midrule
         Latin & 185 & 77.5 & 69.9 \\
         Arabic & 115 & 78.4 & 60.0 \\
         Hebrew & 48 & 87.1 & 70.0 \\
         Sanskrit & 35 & 95.4 & 84.4 \\
         Chinese & 332 & 79.5 & 64.2 \\
         Byzantine & 233 & 79.3 & 68.3 \\
         \bottomrule
     \end{tabular}
     \vspace{-1em}
    \label{tab:per-script}
\end{wraptable}
\begin{figure*}[t!]
    \centering
    \subfloat[Qualitative results for class-agnostic text detection on historical astronomical diagrams.]{%
    \centering
    \includegraphics[width=0.95\textwidth]{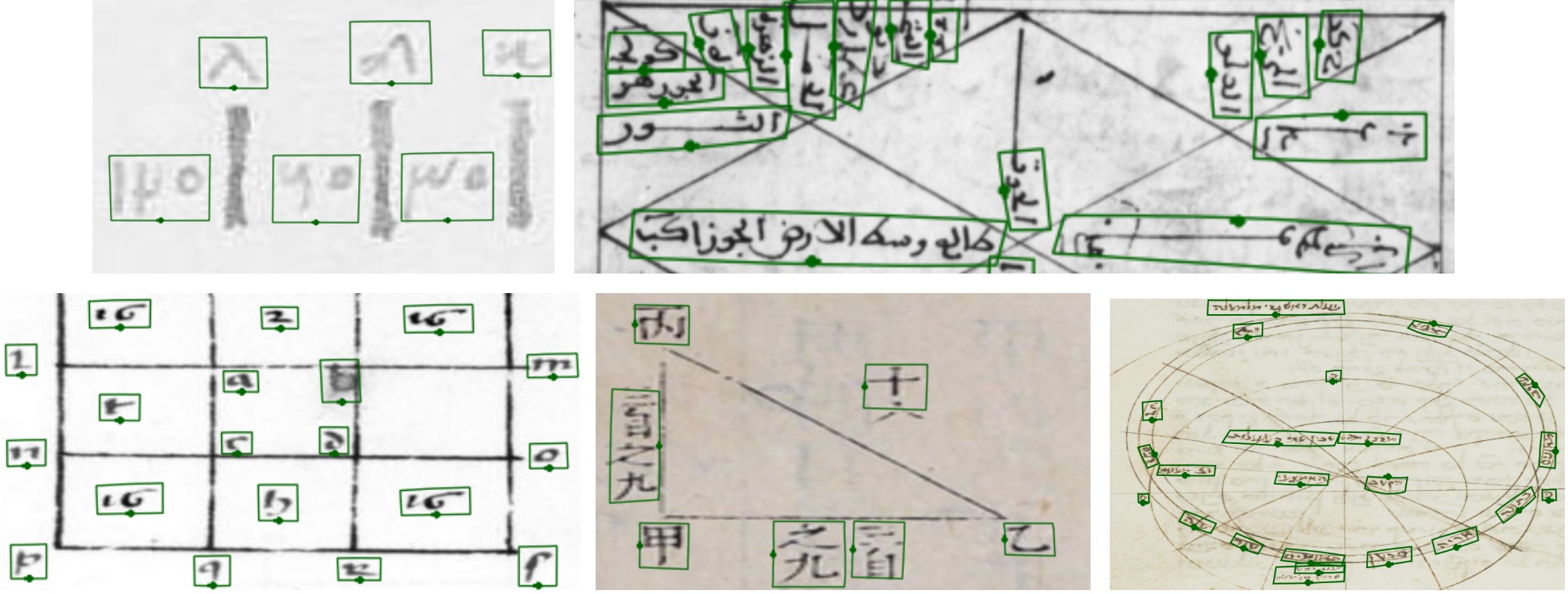}
    }
    \label{fig:qual-diagram}
    \hspace{1em}
    \subfloat[Split text instances.]{%
      \includegraphics[width=0.31\textwidth,height=3cm]{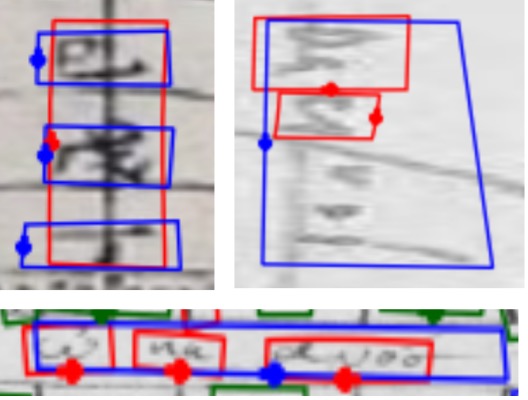}%
    \label{fig:qual-diagramf1}
    }
    \hspace{0.01\textwidth}
    \subfloat[Text around diagrams.]{%
      \includegraphics[width=0.31\textwidth,height=3cm]{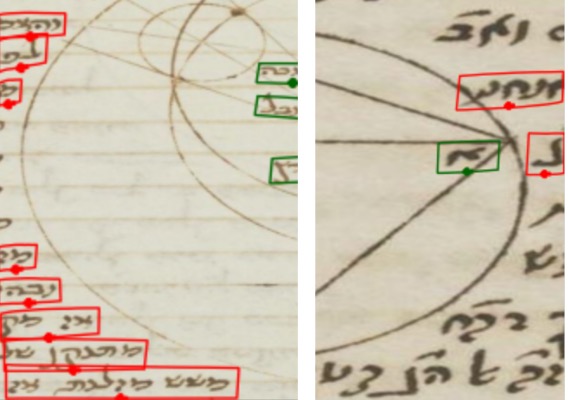}%
    \label{fig:qual-diagramf2}
    }
    \hspace{0.01\textwidth}
    \subfloat[Text-like patterns.]{%
      \includegraphics[width=0.31\textwidth,height=3cm]{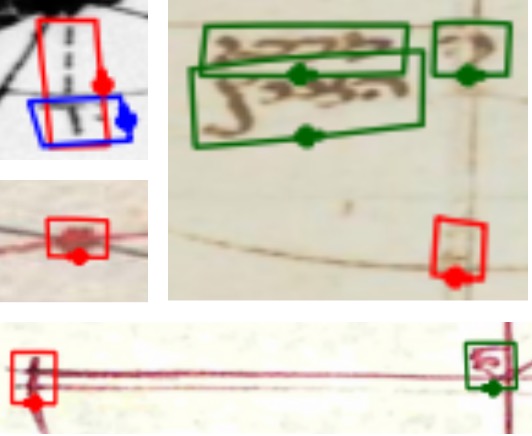}%
    \label{fig:qual-diagramf3}
    }
   \caption{\textbf{Qualitative text detection} on historical astronomical diagrams. (a) Good predictions; (b–d) common error modes. TP, FP, and FN are colored \textcolor{darkgreen}{green}, \textcolor{red}{red}, and \textcolor{blue}{blue}. Failures include (b) text incorrectly split into multiple detections, (c) text outside the diagram that should be ignored, and (d) text-like patterns within drawings mistakenly detected.}
    \label{fig:qual-diagrams}
    \vspace{-2em}
\end{figure*}
Qualitatively, we show that Poly-DETR provides accurate polygons and orientations in diagrams with different styles, contents, and languages. We identify the main failure cases of our approach in~\cref{fig:qual-diagramf1,fig:qual-diagramf2,fig:qual-diagramf3}. Our first failure mode corresponds to incorrect grouping of text elements, mostly a single text block split into multiple detections (\cref{fig:qual-diagramf1}). Our second failure mode is to detect text that is not part of the diagram, which is not annotated but, in some cases, is easy to confuse with diagram text (\cref{fig:qual-diagramf2}). Our third failure mode is to confuse text-like patterns with text (\cref{fig:qual-diagramf3}). We noticed that these failure cases are particularly present for some languages. For example, in Chinese diagrams, the model more often detects drawing elements as text and fails to group individual characters properly. Additionally, a common failure case among all traditions includes marks at the intersection of shapes in diagrams, mistaken for text instances.
\\[5pt]
\SB{\textbf{Per-script analysis.} Our diagrams vary among scripts with language and style. The distribution reflects the inherent scarcity of experts and extant source. Higher performance in Hebrew and Sanskrit, despite fewer samples, stems from their lower structural complexity compared to Arabic, Chinese, or Latin. F1-O shows no clear bias due to the balance between scripts with different reading orders (see~\cref{tab:per-script}).}

\subsection{Class-aware text region detection in astronomical diagrams}
\label{sec:res_cls_diag}
In~\cref{tab:text-class}, we present quantitative results of TESTR~\cite{zhang2022text} and Poly-DETR. For TESTR without finetuning, we associate the prediction with more than one character and more than one word to the {\it word} and {\it long} classes, respectively, and the prediction of characters that were not in our set to the {\it other} class. For TESTR with finetuning, we associate our {\it word}, {\it long}, {\it other}, and {\it symbol} classes with unused characters to finetune the network.
%
%
Finetuning proves critical for both methods
and pretraining provides a clear boost for our method. 
\begin{wraptable}{r}{0.58\textwidth}
    \centering
    \vspace{-2.5em}
    \caption{\textbf{Quantitative results for class-aware text detection} on historical astronomical diagrams. \textit{pt.:} pretraining, \textit{ft.:} finetuning.}
    \vspace{-1em}
    \resizebox{0.55\textwidth}{!}{
    \small
    \begin{tabular}{@{}lcccccc@{}}
    \toprule
         Method &  \textit{pt.} & \textit{ft.}  & mF1 & mAP50 \\
         \midrule
        TESTR~\cite{zhang2022text} & scene text & \checkmark & 69.4 & 68.3 \\          
        ~ w/o finetuning & scene text & & 1.9 & 0.9 \\   
         Poly-DETR & synthetic & \checkmark & 64.4 & 63.4 \\
         ~ w/o finetuning & synthetic & &  6.9 & 5.9 \\
         ~ w/o pretraining & none & \checkmark  & 34.2 & 33.4 \\
         \bottomrule
     \end{tabular}}
     \vspace{-2em}
    \label{tab:text-class}
\end{wraptable}
Again, after finetuning, TESTR clearly outperforms our baseline. We believe this is again mainly due to the large-scale pretraining on scene text datasets, and emphasized by the small scale of the Latin diagrams dataset.
\begin{figure*}[t!]
    \centering
    \vspace{-1em}
    \subfloat[Qualitative results for class-aware text detection on historical astronomical diagrams.]{%
    \centering
    \includegraphics[width=0.8\textwidth]{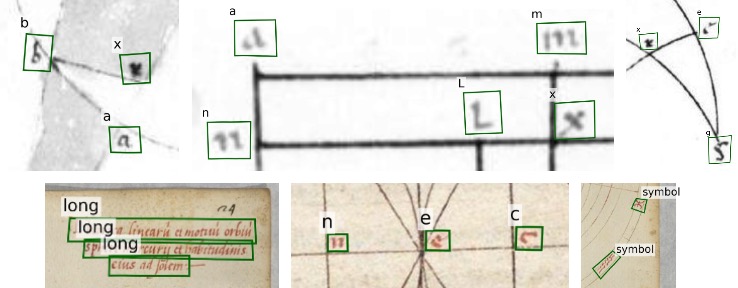}
    } \label{fig:qual-latdiagram}
    \hspace{1em}
    \\
    \subfloat[Splitting text region into sub-categories.]{%
      \includegraphics[width=0.27\textwidth,height=2.5cm]{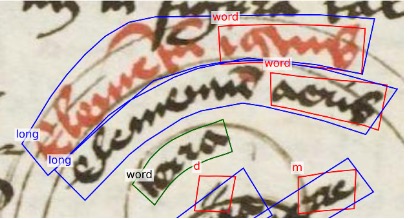}%
    \label{fig:qual-latdiagramf1}
    }
    \hspace{0.01\textwidth}
    \subfloat[Text around diagrams.]{%
      \includegraphics[width=0.40\textwidth,height=2.5cm]{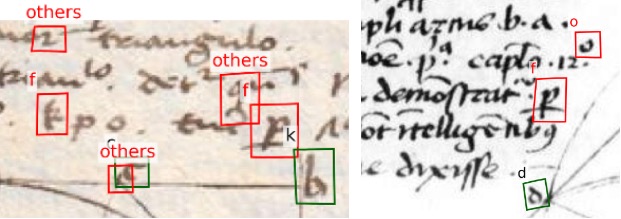}%
    \label{fig:qual-latdiagramf2}
    }
    \hspace{0.01\textwidth}
    \subfloat[Text-like patterns.]{%
      \includegraphics[width=0.23\textwidth,height=2.5cm]{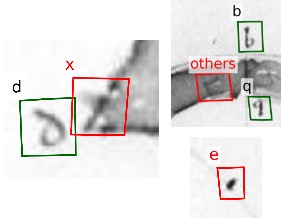}%
    \label{fig:qual-latdiagramf3}
    }
\caption{\textbf{Qualitative results for class-aware text detection} on historical astronomical diagrams. (a) Good predictions; (b–d) common error modes. TP, FP, and FN are colored \textcolor{darkgreen}{green}, \textcolor{red}{red}, and \textcolor{blue}{blue}. Failures: (b) split text regions, (c) text outside the diagram, and (d) ambiguities from inking/text-like patterns.}
    \label{fig:qual-text}
    \vspace{-2em}
\end{figure*}
Qualitative results are reported in~\cref{fig:qual-text} which highlights diverse and meaningful predictions. We visualize the main failure cases in~\cref{fig:qual-latdiagramf1,fig:qual-latdiagramf2,fig:qual-latdiagramf3}. Similarly to class agnostic detection, we can see {\it long} text split into other classes of characters or {\it word} (\cref{fig:qual-latdiagramf1}) and the text around diagrams confused with the text regions of the diagram (\cref{fig:qual-latdiagramf2}). Lastly, due to ambiguities that arise from inking, resolution, or text-like patterns, our model may struggle to detect characters or classify them correctly (\cref{fig:qual-latdiagramf3}).
\section{Conclusion}
\label{sec:conclusion}
We contribute the first dataset for text detection in historical astronomical diagrams, which includes 948 diagrams with 10,940 text regions spanning 10 centuries and 7 cultural traditions. Our annotations capture accurate text boundaries and their reading order through ordered polygons, as well as single-character labels in Latin diagrams. Our dataset offers a solid foundation for advancing research to understand textual content within historical diagrams. We additionally propose a simple baseline, Poly-DETR, that predicts ordered polygons. We show that it improves the state-of-the-art for text line detection on historical manuscripts for Latin and Chinese benchmarks.
It also provides a strong baseline for our dataset, but it is outperformed by TESTR pretrained on large scale scene text data and finetuned on our dataset.
\\[5pt]
\textbf{Acknowledgements.} This work was funded by the projects EIDA (ANR-22-CE38-0014), VHS (ANR-21-CE38-0008) and the ERC grant DISCOVER (No. 101076028). This work was performed using HPC resources from GENCI-IDRIS (AD010614956R1, AD011015222, AD011015415). The authors thank E. Andriani, J. Chen, S. Guessner, D. Manolova, S. Trigg, M. Vlachou Efstathiou, L. Bensabath, and V. Attias for contributions and feedback.

\bibliographystyle{splncs04}
\bibliography{references}
\end{document}